\pdfoutput=1

\DocumentMetadata{}
\documentclass[sigconf]{acmart}

\AtBeginDocument{%
  \providecommand\BibTeX{{%
    Bib\TeX}}}

\copyrightyear{2025}
\acmYear{2025}
\setcopyright{rightsretained}
\acmConference[KDD '25]{Proceedings of the 31st ACM SIGKDD Conference on Knowledge Discovery and Data Mining V.1}{August 3--7, 2025}{Toronto, ON, Canada}
\acmBooktitle{Proceedings of the 31st ACM SIGKDD Conference on Knowledge Discovery and Data Mining V.1 (KDD '25), August 3--7, 2025, Toronto, ON, Canada}
\acmDOI{10.1145/3690624.3709183}
\acmISBN{979-8-4007-1245-6/25/08}

\makeatletter
\gdef\@copyrightpermission{
\begin{minipage}{0.2\columnwidth}
\href{https://creativecommons.org/licenses/by/4.0/}{\includegraphics[width=0.90\textwidth]{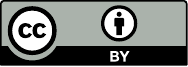}}
\end{minipage}\hfill
\begin{minipage}{0.8\columnwidth}
\href{https://creativecommons.org/licenses/by/4.0/}{This work is licensed under a Creative Commons Attribution International 4.0 License.}
\end{minipage}
\vspace{5pt}
}
\makeatother

\PassOptionsToPackage{usenames,dvipsnames}{xcolor}
\usepackage{algorithmic}
\usepackage{graphicx}
\usepackage{xurl}
\usepackage[usenames,dvipsnames]{xcolor}
\usepackage{algorithm}
\usepackage{tikz}
\usepackage{booktabs}
\usepackage{subcaption}
\usepackage{multirow}
\usepackage{soul}
\usepackage{pifont}
\usepackage{makecell}
\usepackage{float}
\usepackage{threeparttable}
\usepackage{balance}
\usepackage{eucal}
\usepackage{caption}
\usepackage{array}
\usepackage{setspace}
\usepackage{lipsum}
\usepackage{adjustbox,mdframed}
\usepackage{xspace}
\usepackage{marginnote}
\usepackage[algo2e, ruled, linesnumbered]{algorithm2e} 
\usepackage[hang,flushmargin]{footmisc}
\usepackage{eucal}
\DeclareFontFamily{OT1}{mathc}{}
\DeclareFontShape{OT1}{mathc}{m}{n}{ <-> mathc10 }{}

\usepackage{tabularx}
\usepackage[toc,page]{appendix}
\usepackage{hyperref}
\hypersetup{
    colorlinks=true,
    linkcolor=blue,
    filecolor=magenta,      
    urlcolor=cyan,
}

\usepackage{enumitem}

\newcommand{\ra}[1]{\renewcommand{\arraystretch}{#1}}
\newcommand{\model}{\textsc{JEDIS}}
\newcommand{\git}{https://github.com/minkyoo9/JEDIS}

\definecolor{BlueViolet}{RGB}{71,57,146}
\definecolor{Brown}{RGB}{121,37,0}
\definecolor{BrickRed}{RGB}{182,50,28}

\def\BibTeX{{\rm B\kern-.05em{\sc i\kern-.025em b}\kern-.08em
    T\kern-.1667em\lower.7ex\hbox{E}\kern-.125emX}}

\newif\ifdiff

\newcommand{\diff}[1]{\ifdiff{\color{black}\fi#1\ifdiff}\fi}

\difftrue

\begin{document}

\title[Covering Cracks in Content Moderation]{Covering Cracks in Content Moderation: Delexicalized Distant Supervision for Illicit Drug Jargon Detection}

\author{Minkyoo Song}
\affiliation{
  \institution{Korea Advanced Institute of Science \& Technology}
    \city{Daejeon}
  \country{Republic of Korea}}
\email{minkyoo9@kaist.ac.kr}

\author{Eugene Jang}
\affiliation{
  \institution{S2W Inc.}
    \city{Seongnam}
  \country{Republic of Korea}}
\email{genesith@s2w.inc}

\author{Jaehan Kim}
\affiliation{
  \institution{Korea Advanced Institute of Science \& Technology}
    \city{Daejeon}
  \country{Republic of Korea}}
\email{jaehan@kaist.ac.kr}

\author{Seungwon Shin}
\affiliation{
  \institution{Korea Advanced Institute of Science \& Technology}
    \city{Daejeon}
  \country{Republic of Korea}}
\email{claude@kaist.ac.kr}

% \author{Minkyoo Song\textsuperscript{1}\hspace{0.7em} Eujene Jang\textsuperscript{2}\hspace{0.7em} Jaehan Kim\textsuperscript{1}\hspace{0.7em} Seungwon Shin\textsuperscript{1} \\
%         \,\textsuperscript{1}KAIST \texorpdfstring{\; \quad} \,\textsuperscript{2}S2W Inc.\\
%         \,\textsuperscript{1}\texorpdfstring{\texttt{\{minkyoo9, jaehan, claude\}@kaist.ac.kr} \;} 
%         \,\textsuperscript{2}\texttt{\{genesith\}@s2w.inc}
% }


\begin{abstract}
In light of rising drug-related concerns and the increasing role of social media, sales and discussions of illicit drugs have become commonplace online.
Social media platforms hosting user-generated content must therefore perform content moderation, which is a difficult task due to the vast amount of jargon used in drug discussions.
Previous works on drug jargon detection were limited to extracting a list of terms, but these approaches have fundamental problems in practical application.
First, they are trivially evaded using word substitutions. Second, they cannot distinguish whether euphemistic terms (\textit{pot}, \textit{crack}) are being used as drugs or as their benign meanings.
We argue that drug content moderation should be done using contexts, rather than relying on a banlist.
However, manually annotated datasets for training such a task are not only expensive but also prone to becoming obsolete.
We present \model{}, a framework for detecting illicit drug jargon terms by analyzing their contexts.
\model{} utilizes a novel approach that combines distant supervision and delexicalization, which allows \model{} to be trained without human-labeled data while being robust to new terms and euphemisms.
Experiments on two manually annotated datasets show \model{} significantly outperforms state-of-the-art word-based baselines in terms of F1-score and detection coverage in drug jargon detection. We also conduct qualitative analysis that demonstrates \model{} is robust against pitfalls faced by existing approaches.
\end{abstract}





\maketitle

\section{Introduction}
\label{introduction}
\begin{figure*}[t]
    \centering
    \includegraphics[width=0.99\linewidth]{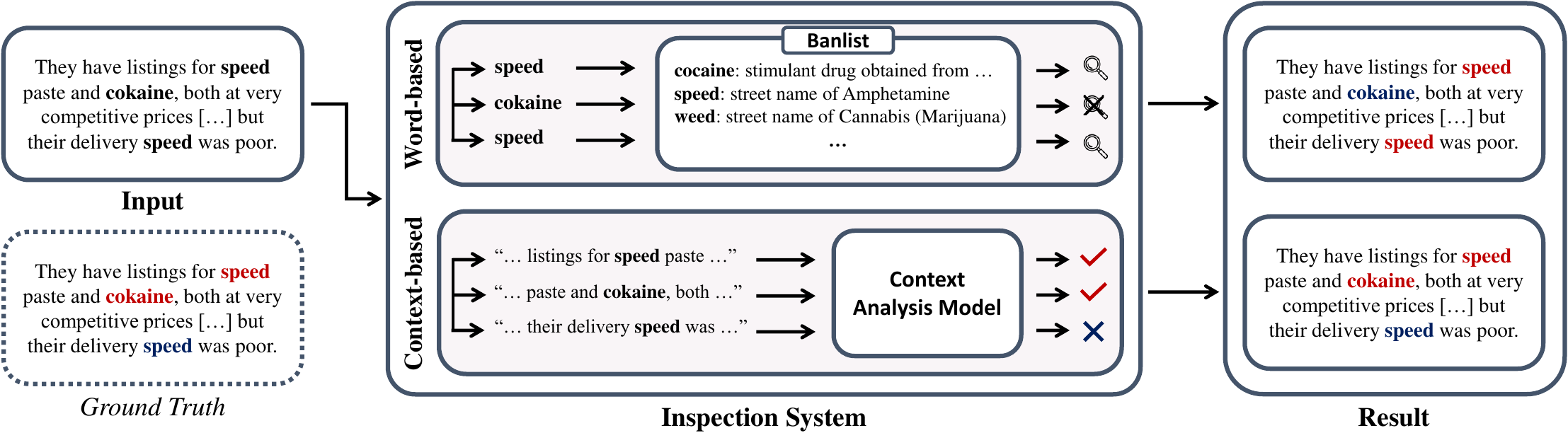}
    \caption{Content moderation procedures of word-based system and context-based system. Bolded words in \textit{Input} were predicted by each system. In the \textit{Result}, words detected as drug jargon are colored \textcolor{BrickRed}{red}, and words predicted as non-drugs are colored \textcolor{BlueViolet}{blue}.}
    \label{fig:motivation}
\end{figure*}

Drug-related deaths are at record highs, with annual drug overdose deaths in the US more than doubling since 2015~\cite{od_deaths, od_deaths2}.
In March 2022, the United Nations urged governments to further regulate social media, citing a link between exposure to social media and drug abuse~\cite{united_nations}.
Illicit sales and discussion of drugs have become prevalent online not only on anonymous marketplaces~\cite{silkroad,dnmArchives} but also on social media~\cite{insta_meta,zhu2021self}.
On top of illegal drugs, online discussion of legal drugs for non-medical recreational usage has also become commonplace~\cite{NPS-forum, nonmedical}.
Without sufficient moderation, social media can expose the general public to sales of illegal drugs~\cite{detect_fent,detect_opioid} and dangerous drug misinformation~\cite{drug_bias}.

Content moderation is a difficult and labor-intensive task.
Although content moderators often face psychological distress from large volumes of explicit content~\cite{moderator_psych}, automating moderation of user-generated content is challenging.
Even for human moderators, understanding the slang and jargon used in illegal content requires significant domain knowledge.
Previous works have addressed this by discovering jargon terms~\cite{klingon, cantreader, china_ug}, interpreting jargon terms~\cite{darkjargon, darkjargon_net}, or both~\cite{zhu2021self}.
Practically, however, moderation systems relying on a banlist of jargon terms suffer from two main limitations.
Figure~\ref{fig:motivation} shows an example input that demonstrates the limitations of a word-based moderation system.
First, word-based restrictions are easily bypassed by alternative terms for moderated words~\cite{hatespeech_aaai}.
This not only includes newly emerging words, but also misspellings of known banned words  (\textit{cokaine}, \textit{c0caine}).
Second, euphemistic jargon has benign usages that can cause false positives.
While \textit{speed} requires moderation when referring to amphetamines, it should not be moderated when referring to quickness.
In order to effectively address such limitations, moderation systems must be trained to evaluate words within their specific context. 
Thus, for more practical content moderation automation, we define \textit{jargon detection} as a \diff{sentence-level} task of identifying if a word, within its specific context, is used as drug jargon.
\diff{To the best of our knowledge, this sentence-level approach for deriving contextual information is a novel direction in drug content moderation research.}

We propose \textbf{\model{}} (Jargon and Euphemism Detection of Illicit Substances), a novel drug jargon detection framework that performs context-aware detection of jargon instances.
\model{} utilizes distant supervision to keep up with complex and fast-evolving drug slang without depending on expensive manual annotations from domain experts.
Drug jargon contexts from an unlabeled text corpus are learned by leveraging only a small list of known drug seed terms.
\diff{
Given an incomplete and small knowledge base for distant supervision, the model may overfit to the seed terms.
To generalize detection, \model{} employs an innovative training strategy that effectively utilizes learned contexts.
Specifically, \model{} uses two independent modules to represent contextual and lexical information separately. The delexicalized context-only module mitigates the excessive focus on the target word by extracting context information alone. The word attribute module preserves target word information to address imprecise detection caused by delexicalization. This allows \model{} to robustly detect unrecognized terms and accurately distinguish between euphemisms.}

% \model{} uses an innovative training strategy to generalize drug jargon detection from the learned contexts without overfitting to the seed terms.
% Specifically, \model{} uses a delexicalized context-only module that explicitly represents contexts alone by masking the target word.
% \diff{Furthermore, to address imprecise detection caused by delexicalization, \model{} independently utilizes target word information through a word attribute module.
% By combining contextual and lexical information from each module, \model{} can robustly detect unrecognized terms and precisely distinguish between euphemisms.}

We construct two human-labeled datasets to evaluate moderation approaches on user-generated drug context sentences.
Our findings suggest that word-based methods perform poorly with content in the wild due to their limitations.
We find that \model{} substantially outperforms state-of-the-art moderation approaches in F1-score across both evaluation datasets.
Specifically, \model{} successfully detects an average of 4.23 times more various drug jargon terms compared to word-based baselines.
It also highlights the need for a tailored context-based method for effective drug jargon detection.
Qualitative analysis shows \model{} detects jargon in scenarios that previous approaches struggled with.

\vspace{0.5cm}
Our contributions are as follows:
% \begin{itemize}[leftmargin=*]
\begin{itemize}
    \item \diff{Propose \model{}, a drug jargon detection framework robust to unseen words and euphemisms by effectively combining contextual and lexical information without requiring labeled training data.}
    \vspace{1mm}
    \item \diff{Propose delexicalized distant supervision for \model{} training, enabling generalization to new terms without overfitting to a small set of drug seed terms.}
    \vspace{1mm}
    \item Create two datasets of forum texts annotated for drug jargon instances to evaluate drug jargon detection\footnote{Our code is available at \href{\git}{\git}.}.
    \vspace{1mm}
    \item Empirically demonstrate limitations of word-based moderation in practical scenarios, which suggest the necessity of context-based moderation.
    \vspace{1mm}
    \item \diff{Empirically show shortcomings of recent language model with prompts in drug jargon detection, underscoring the need for a tailored detection method.}
    % \vspace{0.5mm}
    % \item Identify various factors that contribute to content moderation through an ablation study.
    % \vspace{0.5mm}
    % \item Discuss linguistic challenges involved in automatic drug content moderation.
\end{itemize}

\section{Background}
% Needs to describe the overview here so that the things described don't feel out of place.
% We present a drug jargon detection system that uses distant supervision to find illicit drug jargon using contexts. 
% \textcolor{red}{In this section, we provide background information and related works. We also present the motivation for a context-based content moderation system.}

% \begin{figure*}[t]
%     \centering
%     \includegraphics[width=0.75\linewidth]{Figures/Motivation.pdf}
%     \caption{Content moderation procedures of word-based approaches and context-based approaches. Bolded words in \textit{Input for Inspection} were predicted by each approach. In the \textit{Inspection Result}, words detected as drug jargon are colored \textcolor{BrickRed}{red} and words predicted as non-drugs are colored \textcolor{BlueViolet}{blue}.}
%     \label{fig:motivation}
% \end{figure*}

\subsection{Preliminaries}
\label{preliminaries}
% \begin{figure*}[t]
%     \centering
%     \includegraphics[width=0.99\linewidth]{Figures/Approach_overview.pdf}
%     \caption{Detection Approach Overview}
%     \label{fig:apporach_overview}
% \end{figure*}
% To avoid confusion, we clearly define the terms used in this paper. Note that these terms are often used inconsistently or interchangeably between works.
We define \textbf{drug jargon} as words used to refer to illicit drug substances or abused legal drug substances. This can include both explicit substance names (\textit{MDMA}, \textit{ketamine}) and street names (\textit{Molly}, \textit{ket}).
Some jargon terms are \textbf{euphemisms} (\textit{speed}, \textit{pot}), polysemous words that have alternate innocuous meanings.
It should be noted that slang terms are not necessarily euphemisms (\textit{fent} is slang for \textit{fentanyl} but has no benign alternative usage).
% Similarly, euphemisms are not always used as drug jargon, even in an illegal drug context (the word \textit{speed} in ``[product from dealer] had good \textit{speed} of delivery'' does not refer to \textit{amphetamine}).
We formalize the task of \textbf{drug jargon detection} as a binary classification problem.
\diff{Given a sentence and a word in the sentence, determine whether the word is used as a drug jargon in the specific context.}
This task is contrary to past works, which were typically focused on extracting a list of drug jargon terms from a target corpus.

\begin{figure*}[t]
    \centering
    \includegraphics[width=0.97\linewidth]{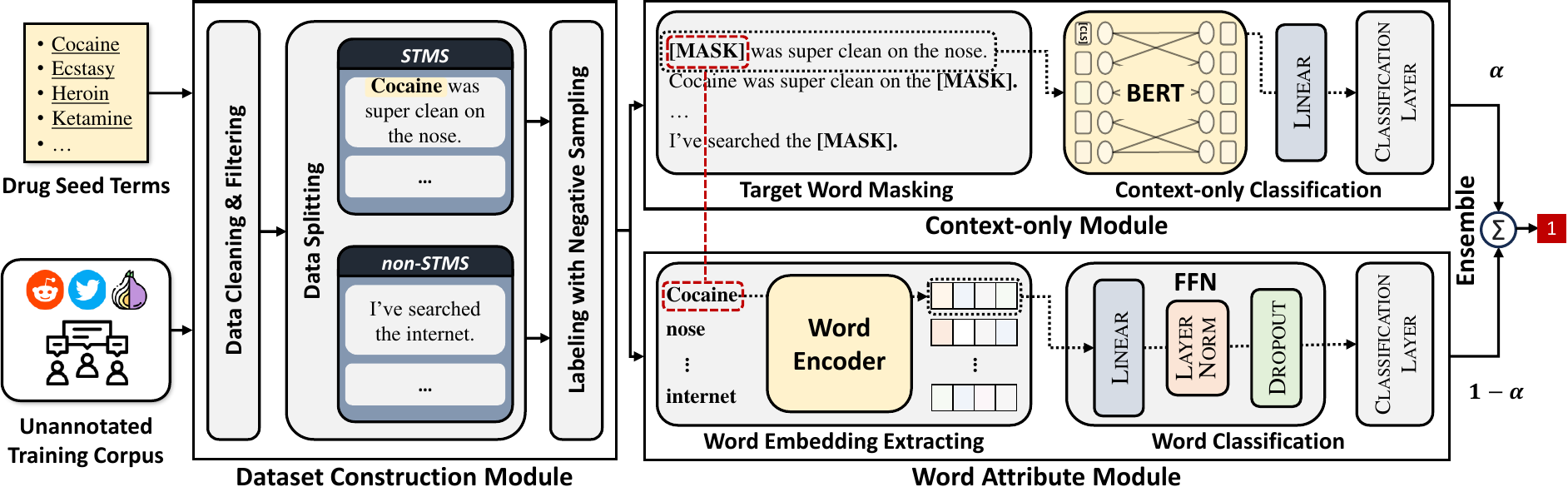}
    \caption{\diff{Overview of \model{}.}}
    \label{fig:overall_workflow}
\end{figure*}

\subsection{Related Work}
\noindent\textbf{Illegal content in social media.}
Detecting abusive content on social media is a challenging task~\cite{SoKModeration}, especially when it is implicitly abusive~\cite{abusive_implicit, risch-etal-2020-offensive}. 
Works on hate speech have identified the usage of euphemisms~\cite{hatespeech_aaai, hatespeech_det} and emojis~\cite{kirk-etal-2022-hatemoji} to circumvent moderation.
Previous works on Twitter data found illegal trafficking of fentanyl~\cite{detect_fent} and opioids~\cite{detect_opioid}.
Previous research on Instagram data identified drug traffickers using text~\cite{insta_text}, multimodal combinations of text and image~\cite{insta_multimodal}, and heterogeneous graphs that incorporate meta-knowledge~\cite{insta_meta}.
These works are limited to finding drug traffickers, and not applicable to drug discussions on social media.
Such discussions are prone to inaccurate and biased information that can lead to fatal results~\cite{drug_bias}.

\vspace{0.1cm}
\noindent\textbf{Jargon detection.}
The languages used by underground communities and marketplaces have been of significant interest to researchers~\cite{keks, coda, jin2023darkbert}.
A common line of jargon detection research follows a set expansion formulation: expanding a set of known drug names by finding similar entities.
Euphemisms of known drugs were found using masked language modeling (MLM) on informative sentences~\cite{euph_mlm,zhu2021self}.
Other works have used a set of known jargon to find new jargon from darknet websites~\cite{chinese_mlm}, underground market groups~\cite{china_ug}, and search engine optimization sites~\cite{klingon}.
Another line of approach is to compare text corpora to identify atypical word usage.
Yuan et al.~\cite{cantreader} leveraged differences in Word2Vec embeddings between different types of corpora to discover words that are used in semantically different ways.
Other works used word probabilities to translate ``dark'' jargon to their ``clean'' counterparts across corpora~\cite{darkjargon,darkjargon_net}.
\diff{However, all these methods primarily generate a list of drug jargon after analyzing a target corpus. 
We contend that such word-based approaches are limited in practical application and advocate for a context-based moderation strategy.}

% \begin{figure*}[t]
%     \centering
%     \includegraphics[width=0.92\linewidth]{Figures/JEDIS_overview.pdf}
%     \caption{\diff{Overview of \model{}.}}
%     \label{fig:overall_workflow}
% \end{figure*}

\vspace{0.1cm}
\noindent\textbf{Distant supervision and delexicalization.}
Large language models (LLMs)~\cite{devlin2018bert} have greatly progressed natural language processing, but such models still require annotated datasets to approach complex tasks such as content moderation.
However, annotated datasets are expensive and prone to become obsolete as the language of the domain changes.
Distant supervision solves the problem of annotation by automatically labeling data using an existing knowledge base. Distant supervision has been utilized for relation extraction~\cite{distant_og, distant_re,quirk-poon-2017-distant} and named entity recognition~\cite{distant_ner, distant_ner2}.

A known challenge in distant supervision is that using an incomplete knowledge base generates noisy labels~\cite{distant-incomplete-kb}.
\diff{Our problem setting is an extreme case where the knowledge base of drug names is very small and incomplete, unlike the Freebase database used for relation extraction containing 116 million relation instances~\cite{distant_og}, or the administrative privileges of the Baidu search engine used for jargon detection~\cite{klingon}. Such extensive resources are often infeasible for moderators.}
To generalize detection to unknown terms instead of simply memorizing known words, we propose a novel approach that integrates delexicalization into the distantly supervised data.
Delexicalization is the concept of replacing the target entities from a text, so that tasks focus on contexts without relying on the target entities.
This technique has been used to improve fact verification~\cite{fact_verif} and spoken language understanding~\cite{spoken}.
\vspace{-0.1cm}
\section{\model{}}
% \begin{figure*}[t]
%     \centering
%     \includegraphics[width=0.92\linewidth]{Figures/JEDIS_overview.pdf}
%     \caption{Overview of \model{}}
%     \label{fig:overall_workflow}
% \end{figure*}
\label{methodology}

% \subsection{Overview}
% \label{overall_workflow}

We present an overview of \model{} in Figure~\ref{fig:overall_workflow}.
In \textit{dataset construction module}, an unannotated training corpus of drug domain texts is cleaned and filtered to maintain the quality of the dataset. 
\diff{The corpus is then split and carefully labeled using drug seed terms and negative sampling to construct the training dataset.}
Then, \model{} uses the dataset to train two modules: the \textit{context-only module} and the \textit{word attribute module}.
In context-only module, \diff{inspected} target words are masked from training sentences.
These masked sentences are then used to train a BERT-based sequence classification model, without exposing the model to the original target word \diff{(i.e., delexicalization)}.
This module is intended to detect drug jargon from contexts alone.
On the other hand, the word attribute module utilizes a vector representation of the target word within its context.
This is intended to utilize lexical information to understand the semantic role that drug seed terms play in drug-related contexts.
To leverage information from both modules, we use an ensemble approach combining both outputs to make final predictions.

\subsection{Dataset Construction Module}
\label{data_processing}
\diff{Since we propose a distant supervision approach for \model{} training, dataset construction does not require annotated data. Instead, it automatically labels unannotated training data using a knowledge base of drug seed terms, enabling \model{} to learn drug contexts.}
\diff{To do this, dataset construction module consists of four steps: data cleaning, filtering, splitting, and labeling with negative sampling.}
In the \textbf{Data Cleaning}, we first utilize the sentence tokenizer from NLTK to separate the training corpus into sentences.
Then, we remove non-ASCII characters, HTML patterns, and words longer than 16 characters to eliminate noise in the text.
In the \textbf{Data Filtering}, we filter out sentences longer than 128 tokens when tokenized by the BERT tokenizer.
We do this because most sentences over 128 tokens are from spam posts without punctuation.
We also filter out sentences shorter than six words, since they have insufficient context.
The number of sentences after filtering in our training corpora can be referred to in Section~\ref{dataset}. 

\vspace{0.1cm}
\noindent\textbf{Data Splitting.}
After cleaning and filtering the training corpus, we use the list of drug seed terms to split it into two pools of sentences: \textit{Seed Term Mentioning Sentences (STMS)} and \textit{sentences without seed terms (non-STMS)}. The former refers to the sentences that mention at least one drug seed term, and the latter refers to the sentences that do not mention any drug seed terms. 

\vspace{0.1cm}
\noindent\textbf{\diff{Labeling with Negative Sampling.}}
% An appropriate mix of positive and negative training data is essential to effectively train the detection model. 
To create \textit{positive} training data, we can simply annotate the seed terms in the STMS.
For example, in the STMS \textit{``\textbf{Cocaine} was super clean on the nose and gave a great feeling.''}, we mark the seed term \textbf{cocaine} as a positive sample.
% By leveraging both the context surrounding the drug seed term and its intrinsic information, the model is trained to make positive predictions when it sees contexts and word information of known drug jargon.
On the other hand, creating \textit{negative} training data requires more careful sampling.
% Data is annotated by selecting an alphanumeric noun (that is not a seed term) as a target word.
We sample negative data from both the STMS and non-STMS sentence pools.
$Neg_{STMS}$ are negative data created by selecting an alphanumeric noun (that is not a seed term) from STMS.
For example, from the previous STMS example sentence, the word \textbf{nose} can be chosen to create the sample \textit{``Cocaine was super clean on the \textbf{nose} and gave a great feeling.''}.
Training with $Neg_{STMS}$ teaches the model to precisely locate which words in a drug-related sentence are drug jargon.
$Neg_{nonSTMS}$ are negative data that represent contexts without known drug terms. 
From non-STMS sentences, any alphanumeric noun (like \textbf{internet}) can be chosen to create negative samples such as \textit{``I've searched the \textbf{internet} and can't find anything too helpful.''}.
% --$Neg_{nonSTMS}$.
% $Neg_{nonSTMS}$ describes contexts without known drug terms.
% Although it is conceptually sufficient to use just $Neg_{nonSTMS}$, we also use $Neg_{STMS}$ to create negative data.

Since the list of known drug terms is highly incomplete, negative labels are potentially noisy.
% Both negative sentence types suffer from this limitation, but to different degrees.
% Since STMS are more likely to be discussions explicitly mentioning drugs, $Neg_{STMS}$ are more prone to noise than $Neg_{nonSTMS}$.
% However, $Neg_{STMS}$ can be more challenging, as it teaches the model to precisely locate which words in a drug-related sentence are drug jargon.
Empirically, we find that controlling the ratio of negative training data is important to maximize performance in jargon detection.
Therefore, we use two hyperparameters to control negative sampling: $\mathcal{R}_{STMS}$ and $\mathcal{R}_{nonSTMS}$.
% $\mathcal{R}_{STMS}$ controls the influence of $Neg_{STMS}$.
$\mathcal{R}_{STMS}$ is the maximum number of negative samples extracted from a single STMS sentence (if the sentence contains less than $\mathcal{R}_{STMS}$ alphanumeric nouns, all are used).
% $\mathcal{R}_{nonSTMS}$ controls the influence of $Neg_{nonSTMS}$.
$\mathcal{R}_{nonSTMS}$ represents the ratio of non-STMS sentences selected per STMS sentence.
For each selected non-STMS sentence, a single negative sample is made by choosing a random alphanumeric noun as the target.
The impact of the two types of negative training data can be regulated by the two hyperparameters, which we can optimize for performance.
%We optimize the hyperparameters for performance in jargon detection.
Further discussion about the negative ratios is included in Section~\ref{detection_evaluation} and ~\ref{further_exps}.

\subsection{Context-only Module}
\label{context_learning}
\noindent\textbf{Target Word Masking.}
Because we construct our training dataset by explicitly utilizing the drug seed terms, directly training on this data can cause a classification model to heavily overfit to the seed terms by simply memorizing them.
This is especially the case since our seed term list is relatively small (46) compared to other knowledge bases used for distant supervision.
To minimize the effect of seed terms, we \diff{\textit{delexicalize}} the prediction process by masking the target word (annotated before) with the [MASK] token.
Therefore, the classification model must predict whether a target word is drug jargon or not without actually seeing the word in question.
This setting removes the lexical influences of the inspected words while emphasizing the contexts.

\vspace{0.1cm}
\noindent\textbf{Context-only Classification.}
% As mentioned in Section~\ref{overall_workflow}, we designed a context-only learning step to investigate if a masked word is used for drug jargon in the sentence by conducting the sequence classification task.
The classifier takes masked sentences and determines if the masked word from a sentence is a drug jargon.
To do this, we construct a sequence classification model using BERT~\cite{devlin2018bert}.
First, the BERT (\textit{bert-base-uncased}) model is adapted to the target domain through a pretraining process.
This has been shown to improve performance, especially on domains where the linguistic characteristics deviate from conventional contexts~\cite{dapt}.
% We take the \textit{bert-base-uncased} model and pretrain on the MLM task on the entire target corpus.

Then, sentences from the constructed dataset are used to train the model.
Each masked training sentence is surrounded by the [CLS] token and the [SEP] token and inputted into the BERT model.
The BERT pooled output $\textbf{\large{o}}_{pooled}$ is extracted by passing the [CLS] token embedding $\textbf{h}_{[CLS]}$ from the last hidden layer of BERT through the linear layer (weights $\textbf{W}_{1} \in \mathbb{R}^{dim \times dim}$ and bias $\textbf{b}_{1}\in \mathbb{R}^{dim}$, $dim$ represents the dimension of BERT embedding) then applying \textit{tanh}.
\begin{align}
    \textbf{\large{o}}_{pooled} = \textit{tanh}(\textbf{W}_{1}\textbf{h}_{[CLS]}+\textbf{b}_{1})
\end{align}
It represents the context of the whole sentence including the [MASK] token.
Finally, we apply classification layer $\textbf{W}_{2} \in \mathbb{R}^{dim \times 1}$ and \textit{sigmoid} on the pooled output to conduct a binary classification task by calculating the probability that the masked word is drug jargon.
\begin{align}
    \textit{\large {$p$}}_{\large c} =
    \textit{sigmoid}(\textbf{W}_{2}\textbf{\large{o}}_{pooled})
\end{align}
% This probability output is later ensembled as described in Section~\ref{ensemble}.

% \mk{
% The pooled output is then fed into the classification layer $\textbf{W}_{2} \in \mathbb{R}^{dim \times 1}$ to obtain the final logit of the context-only representation.}
% \begin{align}
%     \textit{{${{logit}}_{context}$}} =
%     \textbf{W}_{2}\textbf{\large{o}}_{pooled}
% \end{align}

\subsection{Word Attribute Module}
\label{word_learning}
\noindent\textbf{Word Embedding Extracting.}
Relying only on context without seeing the target word may cause false positives\footnote{We further discuss the cases of false positives associated with linguistic challenges in Section~\ref{challenges}.}.
To also utilize lexical information and mitigate false positives, we utilize a word encoder to obtain embeddings of the target word from each sentence.
Crucially, the encoder must generate \textit{contextual embeddings}, creating the appropriate representations depending on word context. 
% Without this capability, our framework simply memorizes static embeddings from the drug seed term list, potentially leading to overfitting.
Thus, we utilize a BERT model pretrained on the target domain as the word encoder.
This allows us to utilize the context-sensitive intrinsic information of target words to make jargon detection more precise.
This word encoder operates independently from the BERT of the context-only module.
The model weights are frozen to prevent overfitting caused by seeing the complete sentences through the entire framework.

\vspace{0.1cm}
\noindent\textbf{Word Classification.}
We train a feed-forward network and a classification layer that use the dynamic embeddings of the target word.
Each word embedding $\textbf{\large{e}}_{word}$ is passed through a feed-forward network consisting of a linear layer (weights $\textbf{W}_{3} \in \mathbb{R}^{dim \times dim}$ and bias $\textbf{b}_{3} \in \mathbb{R}^{dim}$), $LayerNorm$, \textit{tanh}, and $Dropout$.
\begin{gather}
\textbf{z}_{word} = \textbf{W}_{3}\textbf{\large{e}}_{word} + \textbf{b}_{3}\\
\textbf{\large{o}}_{{word}} = \textit{Dropout}(\textit{tanh}(\textit{LayerNorm}(\textbf{z}_{word})))
\end{gather}
By doing this, information of the target word within its context is represented as $\textbf{\large{o}}_{{word}}$.
This enhanced representation is then used as the input for the classification layer $\textbf{W}_{4} \in \mathbb{R}^{dim \times 1}$ and \textit{sigmoid} function to obtain the probability that the target word is drug jargon.
\begin{align}
    \textit{\large {$p$}}_{\large w} =
    \textit{sigmoid}(\textbf{W}_{4}\textbf{\large{o}}_{word})
\end{align}
% This probability output is later ensembled as described in Section~\ref{ensemble}.

\subsection{Ensemble Training and Prediction}
\label{ensemble}
The two probabilities obtained from the context-only and word-attribute modules are used to generate the final prediction of the jargon detection task.
In the training phase, probability $\textit{\large {$p$}}_{\large c}$ is used to calculate the loss for the context-only classification model by evaluating the binary cross-entropy loss between the predicted probability $\textit{\large {$p$}}_{\large c}$ for the training sample $s$ in training dataset $\mathcal {D}$ and the ground truth label $y \in \{0,1\}$.
\begin{align}
    \mathcal{L}_{\large c} = -\frac{1}{\lvert \mathcal{D} \rvert} \sum_{(s_{i},y_{i}) \in \mathcal{D}} y_{i}\,log\,\textit{\large {$p$}}_{\large c,s_{i}} + (1 - y_{i})\,log\,(1 - \textit{\large {$p$}}_{\large c,s_{i}})
\end{align}
Similarly, we compute the loss for the word classification model by assessing the binary cross-entropy loss between the predicted probability $\textit{\large {$p$}}_{\large w}$ for the training sample and ground truth label.
\begin{align}
    \mathcal{L}_{\large w} = -\frac{1}{\lvert \mathcal{D} \rvert} \sum_{(s_{i},y_{i}) \in \mathcal{D}} y_{i}\,log\,\textit{\large {$p$}}_{\large w,s_{i}} + (1 - y_{i})\,log\,(1 - \textit{\large {$p$}}_{\large w,s_{i}})
\end{align}
Then, we incorporate a learnable parameter, $\alpha$, to compute the final loss as:
\begin{align}
    \mathcal{L} = \alpha*\mathcal{L}_{\large c} + (1-\alpha)*\mathcal{L}_{\large w}
\end{align}
The classification models within \model{} are trained by minimizing this ensemble loss.
\diff{Thus, \model{} can leverage both contextual and word information with appropriate weights.}
In the prediction phase, we determine the final probability of the target word being drug jargon by weighting the two probabilities, $\textit{\large {$p$}}_{\large c}$ and $\textit{\large {$p$}}_{\large w}$, using the previously described parameter $\alpha$.
\begin{align}
    \textit{\large {$p$}} = \alpha*\textit{\large {$p$}}_{\large c} + (1-\alpha)*\textit{\large {$p$}}_{\large w}
\end{align}
\section{Datasets and Annotation}

\subsection{Datasets}
\label{dataset}
\begin{table}[t]
    \centering
    \caption{Summary of data statistics.}
    \footnotesize
    % \resizebox{\linewidth}{!}{%
        \begin{tabular} {l c c}
            \toprule
                 &\textbf{Reddit Drug}      &\textbf{Silk Road Forum}  \\ \midrule
            \textbf{Posts}     & 1,271,907   & 1,227,847 \\
            \textbf{Sentences}   & 6,793,530    & 3,952,297\\       
            \textbf{Sentences (filtered)}   & 4,799,512   & 3,057,758\\
            \textbf{STMS (\%)} &  281,104 (5.86\%) & 159,944 (5.23\%)\\
            %\textbf{Duration (month / year)}   & 2/2008 - 12/2017  &   6/2011 - 10/2013  \\ \bottomrule      
            % \textbf{Collection period}   & 2/2008 - 12/2017  &   6/2011 - 10/2013  \\ \bottomrule      
            \textbf{Collection period}   & Feb 2008 - Dec 2017  &  June 2011 - Oct 2013  \\ \bottomrule      
                       
        \end{tabular}%
    % }
    \label{tab:data_statistics}
\end{table}
\vspace{0.1cm}
\noindent\textbf{Drug seed terms.}
Since we are using the list of drug seed terms for distant supervision, we wish to find explicit drug terms that are unambiguously used as drugs.
We refer to the drug slang report published by the Drug Enforcement Administration (DEA)~\cite{dea_list}, and take only the explicit drug names to create a list of 46 explicit drug terms\footnote{Illegality may vary depending on local legislation.}. 
% The generated list of drug seed terms is available in Table~\ref{tab:drug_seed_terms} in Appendix~\ref{appendix}.
The generated list is available in our \href{\git}{repository}.

\vspace{0.1cm}
\noindent\textbf{Drug forum corpora.}
We use text data from drug-related forums to train \model{} on user-generated drug discussions. For effective detection, it's essential that our data includes both drug and non-drug contexts. To assess \model{}'s versatility across settings, we employ two different drug-related forum corpora.

The \textit{Reddit Drug corpus}, provided by Zhu et al.\cite{zhu2021self}, is sourced from the public Reddit archive\cite{reddit_query} on Google BigQuery\footnote{https://console.cloud.google.com/bigquery?project=fh-bigquery}. It comprises posts from 46 subreddits related to drugs and darknet markets, such as `Drugs', `DarkNetMarkets', and `SilkRoad', many of which are now inaccessible due to Reddit's policy change~\cite{reddit_ban}. Some subreddits, like `Bitcoin' and `TOR', are only tangentially drug-related but are included to ensure a variety of non-drug contexts. This corpus represents surface web drug discussions.

The \textit{Silk Road Forum corpus}, derived from the darknet market archive by Branwen et al.~\cite{dnmArchives}, represents discussions from the Silk Road forum. Silk Road, once the largest drug-related darknet market, was shut down by the FBI in 2013~\cite{cnn_sr_shutdown}. Its associated forum hosted categories ranging from `Legal' and `Drug safety' to broader subjects like `Security' and `Philosophy'. We incorporate posts from all these categories to ensure a mix of drug and non-drug contexts, representing dark web drug discussions. 
The statistics of our corpora are summarized in Table~\ref{tab:data_statistics}. 
% We also provide a deeper analysis of our corpora, especially regarding the contextual distinctiveness of drug jargon, in Appendix~\ref{data_analysis}.

\subsection{Annotation}
\label{annotation}
To evaluate \model{} quantitatively, we constructed evaluation datasets by annotating \diff{1,100} randomly selected sentences from each forum corpus, excluding sentences with drug seed terms.
\diff{Two annotators from a cybersecurity company specializing in monitoring malicious online activities were recruited to annotate the sentences for drug jargon.}
% Two expert annotators identified words used as drug jargon in the given sentences.
For quality assurance, we conducted a pilot annotation on the first 100 sentences from each corpus, subsequently constructing annotation guidelines. These pilot sentences were excluded from the final evaluation datasets.

Through this manual annotation, we generated two evaluation datasets, each with \diff{1,000} sentences. Table~\ref{tab:annotation_statistics} provides detailed statistics of the annotation results. \diff{The inter-annotator agreement registered Cohen's Kappa values of 0.86 and 0.87 for the respective datasets, indicating almost perfect agreement ($>$ 0.8). }
Both annotation guidelines and the ethical considerations for the dataset construction can be found in Appendix~\ref{appendix} and ~\ref{ethical}, respectively.
\begin{table}[t]
    \centering
    \caption{\diff{Statistics of the annotation results. $^{*}$Only alphanumeric nouns are considered.}}
    \footnotesize
    % \resizebox{\linewidth}{!}{%
        \begin{tabular} {l c c}
            \toprule
                 &\textbf{Reddit Drug}      &\textbf{Silk Road Forum}  \\ \midrule
            \textbf{Sentences}     & \diff{1,000}   & \diff{1,000} \\
            \textbf{Cohen's Kappa}  & \diff{0.86}  & \diff{0.87} \\  
            \textbf{Nouns$^{*}$}   & \diff{4,559}    & \diff{5,531} \\       
            \hspace{0.3cm}\textbf{Drug jargon (\%)}   & \diff{447 (9.8\%)}   & \diff{473 (8.6\%)} \\
            % \hspace{0.3cm}\textbf{Not drug jargon (\%)}   & 1,965 (86.1\%)  &  2,558 (89.6\%) \\ 
            \hspace{0.3cm}\textbf{Unique drug jargon}   & \diff{178}    & \diff{169} \\
            \bottomrule     
        \end{tabular}%
    % }
    \label{tab:annotation_statistics}
\end{table}

\diff{Additionally, in Appendix~\ref{potential_sub_bias}, we discuss how to address potential subjectivity and bias in the annotation process.}
\section{Evaluation}
\label{detection_evaluation}

% We evaluate the effectiveness of \model{} on drug jargon detection for content moderation by utilizing the evaluation datasets constructed in Section~\ref{annotation}.
% We also investigate the robustness and coverage of \model{} with qualitative analysis.
\begin{table*}[t]
    \centering
    \caption{{Evaluation results of drug jargon detection task. Boldface represents the best performance and underline represents the second-best performance. The \textit{\# Jargon} column represents the number of \textit{unique} jargon terms detected by each approach.}}
    \footnotesize
    \begin{tabular} {l c @{ } c @{ } c @{ } c c c @{ } c @{ } c @{ } c}
        \toprule
        & \multicolumn{4}{c}{\textbf{Reddit Drug}}  &  & \multicolumn{4}{c}{\textbf{Silk Road Forum}} \\ \cmidrule{2-5}\cmidrule{7-10}
        & \multicolumn{1}{c}{\textbf{Precision}}  & \multicolumn{1}{c}{\textbf{Recall}} & \multicolumn{1}{c}{\textbf{F1-score}} & \multicolumn{1}{c}{\textbf{\# Jargon}} & & \multicolumn{1}{c}{\textbf{Precision}}  & \multicolumn{1}{c}{\textbf{Recall}} & \multicolumn{1}{c}{\textbf{F1-score}} & \multicolumn{1}{c}{\textbf{\# Jargon}} \\
        \midrule
        % \textbf{Word2Vec (SG)} & \underline{0.8588} &  0.2303 & 0.3632 &  26 & & \textbf{0.8934} &  0.3670 & 0.5203 &  20 \\
        \textbf{Word2Vec~\cite{mikolov2013efficient}} & {\textbf{0.8908}} & {0.2371} & {0.3746}  & {32} & & {\textbf{0.7908}}  & {0.3277}  & {0.4634}  & {39} \\
        \textbf{CantReader~\cite{cantreader}} & {\underline{0.7778}}  & {0.1879} & {0.3027} & {17} & & {\underline{0.7208}}  & {0.2347} & {0.3541} & {22} \\
        \textbf{Zhu et al.~\cite{zhu2021self}} & {0.5897} & {0.2573}  & {0.3583} & {17} & & {0.5026} & {0.4123} & {0.4530} & {27} \\
        \midrule
        \textbf{{PETD~\cite{lee-etal-2022-searching}}} & {0.4760} & {0.2215} & {0.3023} & {28} & & {0.5789} & {0.2326} & {0.3318} & {26}\\
        \textbf{MLM (\textit{w/o pretrain})} & {0.5094} & {0.3043} & {0.3810} & {61} & & {0.4569} & {0.2579} & {0.3297} & {57}\\
        \textbf{MLM} & {0.5241} & {0.6085} & {0.5631} & {98} & & {0.5783} & {0.6321} & {0.6040} & {102}\\
        \textbf{{DarkBERT~\cite{jin2023darkbert}}} & {0.5460} & {0.5705} & {0.5580} & {92} & & {0.4790} & {0.5793} & {0.5244} & {109}\\
        % \textbf{\diff{GPT3.5-Turbo~\cite{gpt3.5turbo}}} & \diff{0.2659} & \diff{\textbf{0.6913}} & \diff{0.3841} & \diff{\textbf{124}} & & \diff{0.2315} & \diff{0.6406} & \diff{0.3401} & \diff{\textbf{124}}\\
        \textbf{{GPT4o-mini~\cite{gpt4omini}}} & {0.4307} & {\textbf{0.7919}} & {0.5579} & {\textbf{129}} & & {0.3920} & \textbf{0.7907} & {0.5242} & {\textbf{137}}\\
        \midrule
        \textbf{\model{} (\textit{w/o} $Neg_{STMS}$)} & {0.6052} & {\underline{0.6309}} & {0.6177} & {\underline{105}} & & {0.5367} & {0.6808} & {0.6002} & {112} \\
        \textbf{\model{} (\textit{w/o pretrain\&word})} & {0.6318} & {0.5951} & {0.6129} & {97} & & {0.5368} & {0.6321} & {0.5806}  & {104} \\
        \textbf{\model{} (\textit{w/o word})} & {0.6454} & {0.6107} & {\underline{0.6276}} & {97} & & {0.5507} & {\underline{0.6998}} & {\underline{0.6164}}  & {\underline{116}}\\
        \textbf{\model{} (\textit{w/o pretrain})} & {0.6475} & {0.5794} & {0.6116} & {92} & & {0.5573} & {0.6068} & {0.5810} & {101}\\
        \midrule
        \textbf{\model{}} & {0.6659} & {0.6197} & {\textbf{0.6419}}  & {101} & & {0.5805} & {{0.6934}} & {\textbf{0.6320}} & {113} \\
        \bottomrule
    \end{tabular}%
    \label{tab:evaluation1}
\end{table*}

\subsection{Experimental Setup}
\label{experimental_setup}
% For each alphanumeric noun in the evaluation datasets, models predict whether the word was used as drug jargon in the sentence.

\vspace{0.1cm}
\noindent\textbf{Training \model{}.}
For each corpus, we create a training dataset using empirically selected hyperparameters $\mathcal{R}_{STMS}=5$ and $\mathcal{R}_{nonSTMS}=2$\footnote{Detailed discussion of these hyperparameters will be in Section~\ref{detection_results} and ~\ref{further_exps}.}.
We also limit the range of learnable parameter $\alpha$ as [0.9, 0.99] to emphasize the delexicalized context-only module\footnote{We empirically find that a low value of $\alpha$ leads to \textbf{significant overfitting}, resulting a low recall.}.
% For each dataset, we split the data into train/valid sets at an 8:2 ratio.
% Sentences included in the evaluation dataset were excluded from the training dataset.
% More detailed experimental settings are described in Appendix~\ref{exp_settings}.
\diff{Specifically, we train \model{} using the \textit{unlabeled} Reddit Drug corpus to evaluate the Reddit Drug annotated dataset.
The training data included \textit{329,573} positive instances from STMS, \textit{973,447} negative instances from STMS ($Neg_{STMS}$), and \textit{659,146} negative instances from non-STMS ($Neg_{nonSTMS}$).
For evaluation on the Silk Road Forum annotated dataset, we train the classifier with the \textit{unlabeled} Silk Road Forum corpus.
The training data included \textit{193,167} positive instances, \textit{615,996} $Neg_{STMS}$, and \textit{386,334} $Neg_{nonSTMS}$.
For early stopping, we split the data further into train/valid sets with an 8:2 ratio.}
More detailed experimental settings are described in Appendix~\ref{exp_settings}.

% \footnote{Detailed experimental settings are described in Appendix~\ref{exp_settings}.}
% \input{Tables/evaluation_table_content_moderation}

\vspace{0.1cm}
\noindent\textbf{Baselines.}
\diff{
To ensure fair evaluations, we selected baselines that do not necessitate an annotated dataset for training. 
We examine three word-based jargon detection methods (\textbf{Word2Vec}~\cite{mikolov2013efficient}, \textbf{CantReader}~\cite{cantreader}, and \textbf{Zhu et al.}~\cite{zhu2021self}) as baselines to evaluate their effectiveness in moderating real user-generated conversations.
Unlike JEDIS, these methods generate a list of drug jargon after performing analysis on the \textit{corpus-level}.
Therefore, the models are trained on the entire corpus.
For evaluation, a target word is marked as drug jargon if it appears in the jargon list.
Additionally, we present five context-based baselines (\textbf{PETD}~\cite{lee-etal-2022-searching}, \textbf{MLM \textit{(w/o pretrain})},
\textbf{MLM}, \textbf{DarkBERT}~\cite{jin2023darkbert}, and \textbf{GPT4o-mini}~\cite{gpt4omini}) that inspect \textit{sentence-level} context to determine if a word is used as drug jargon. 
We also evaluate \textbf{variations of JEDIS} to investigate the importance of each component in drug jargon detection. 
Detailed explanations of baselines are described in Appendix~\ref{baselines}}

\subsection{Drug Jargon Detection Results}
\label{detection_results}
\noindent\textbf{Word-based vs Context-based detection.}
The results of our jargon detection experiments are summarized in Table~\ref{tab:evaluation1}. In both datasets, \model{} achieved the highest F1-scores.
The word-based baselines generally have high precision but suffer from low recall.
The context-based approaches scored higher F1-scores overall, suggesting they achieved a better tradeoff in finding drug jargon.

Word-based approaches performed better on the Silk Road Forum dataset than on the Reddit Drug dataset, suggesting that the two datasets might be different in the distribution of more difficult jargon terms (such as euphemisms).
Word-based approaches are expected to have high precision since many recognized drug terms are used only to refer to drugs.
However, their low recall scores strongly suggest that many jargon terms can still evade detection from such systems.
The two euphemism-aware baselines, CantReader and Zhu et al.'s model, achieved differing results.
Compared to other word-based models, CantReader tends to have low recall, while Zhu et al.'s approach sacrifices precision for better recall.
The experiments suggest Zhu et al.'s approach is able to detect more drug contexts, but has more false positives.
The word-based baselines achieved fairly high precision scores, but euphemisms prevent these methods from reaching perfect precision.
The effect of euphemisms in detection is further discussed in Section~\ref{qaulitative_evaluation}, where we further investigate the words selected by the word-based baselines. 

The context-based methods that explicitly consider contexts generally performed better at recall and F1-scores.
\diff{The one exception is PETD, which was specifically designed to detect euphemisms only. 
Its limited performance, especially in recall, highlights the necessity for more precise and comprehensive detection method for effective drug content moderation.
}
% The off-the-shelf MLM baseline outperformed word-based models in the Reddit dataset, but was outperformed by the other context-based models.
Further pretraining the MLM model on the corpora (MLM \textit{(w/o pretrain)} vs MLM) improved performance by a considerable amount.
This suggests that language models require further pretraining on the domain of drug discussions in order to adapt to the lexical and stylistic characteristics of the domain.
This finding is consistent with trends of state-of-the-art models struggling with user-generated texts.
\diff{
However, in the case of DarkBERT, despite not being pretrained on the target corpora, its knowledge of illegal terms obtained from extensive dark web data enabled its detection performance to be comparable to the MLM baseline, particularly in the Reddit Drug dataset. 
A notable observation is that GPT4o-mini with detailed prompts yields high recall and detection coverage.
However, its indiscriminate detection with substantially low precision indicates that even with specifically designed prompts and guidelines (Figure~\ref{fig:prompts}), there is still a need for a method carefully tailored for drug jargon detection.
Although some context-based baselines achieve high performance, their results are not as precise and generalizable as our delexicalized approach. 
\model{} improved performance with its specialized architecture that ensembles the knowledge from both contexts and words.}

\diff{In terms of detection coverage, \model{} exhibited a significant capability in detecting a variety of drug jargon terms, even with satisfactory precision. Specifically, among 178 and 169 unique jargon terms from each dataset, \model{} successfully detected an average of 4.59 times (101 vs 22) more unique terms in the Reddit Drug dataset and 3.86 times (113 vs 29.3) more in the Silk Road Forum dataset compared to the word-based baselines. This underscores \model{}'s superior detection coverage, which significantly exceeds that of existing word-based moderation.}

% \newcolumntype{P}[1]{>{\centering\arraybackslash}p{#1}}
\begin{table*}[t]
    \centering
    \caption{Error analysis of each word-based approach. False positives are colored \textcolor{BrickRed}{red} and false negatives are colored \textcolor{BlueViolet}{blue}.}
    % \footnotesize
    \resizebox{.82\linewidth}{!}{%
    {\small
        \begin{tabularx}{\linewidth}{p{0.13\linewidth} p{0.36\linewidth} p{0.25\linewidth}  p{0.2\linewidth}}
            \toprule
            {\normalsize \textbf{}} &  {\normalsize \textbf{Example sentences}} & {\normalsize \textbf{Limitations}} & {\normalsize \textbf{Example words}}\\ 
            \midrule
            \textbf{Word2Vec~\cite{mikolov2013efficient}} &
            \begin{minipage}{.8\linewidth}
                \includegraphics[width=.95\linewidth]{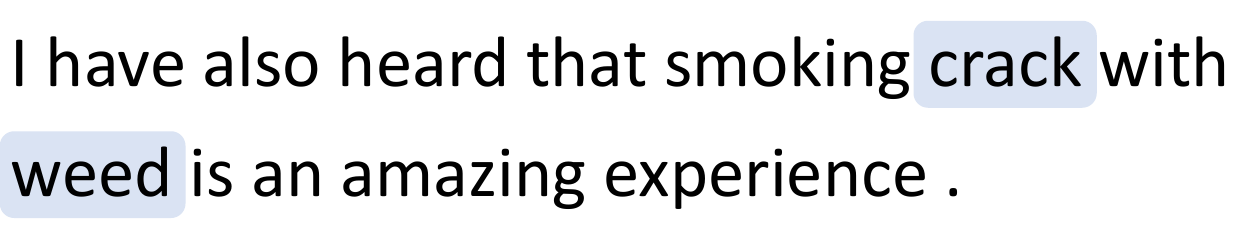}
            \end{minipage} & {\normalsize Cannot find euphemisms} & {\normalsize \textcolor{BlueViolet}{crack}, \textcolor{BlueViolet}{speed}, \textcolor{BlueViolet}{weed}} \\
            \midrule
            \textbf{CantReader~\cite{cantreader}} &
            \begin{minipage}{.8\linewidth}
                \includegraphics[width=.95\linewidth]{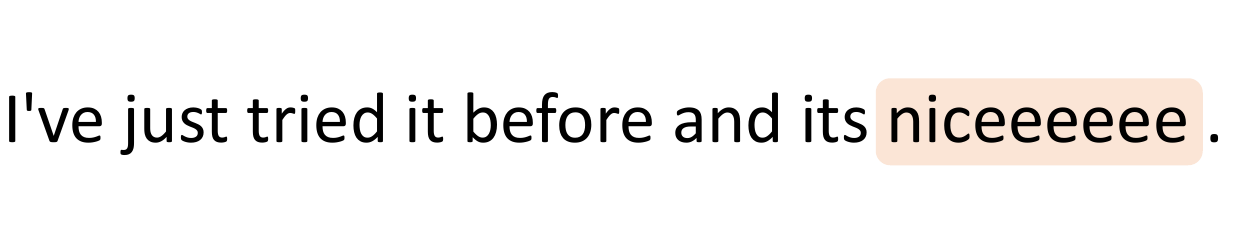}
            \end{minipage} & {\normalsize Sensitive to typos} & {\normalsize \textcolor{BrickRed}{comfused}, \textcolor{BrickRed}{niceeeeee}} \\
            \midrule
            \textbf{Zhu et al.~\cite{zhu2021self}} &
            \begin{minipage}{.8\linewidth}
                \includegraphics[width=.95\linewidth]{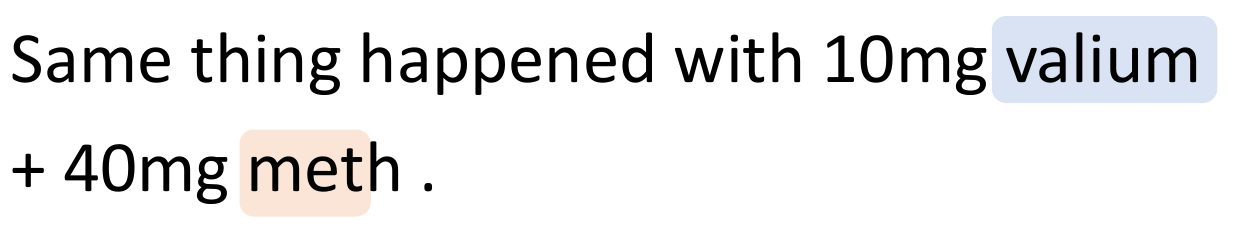}
            \end{minipage} & {\normalsize limited to BERT vocabulary} & {\normalsize \textcolor{BrickRed}{add}, \textcolor{BrickRed}{met}, \textcolor{BlueViolet}{valium}} \\
            % \midrule
            % \textbf{Ours} &
            % \begin{minipage}{.99\linewidth}
            %     \includegraphics[width=.95\linewidth]{Figures/Q_Ours.pdf}
            % \end{minipage} & Insensitive to benignity of words & \textcolor{BrickRed}{alcohol}, \textcolor{BrickRed}{product}, \textcolor{BrickRed}{stuff} \\
            \bottomrule
        \end{tabularx}
    }
    }

    \label{tab:qaulitative_examples}
\end{table*}

\vspace{1mm}
\noindent\textbf{\model{} ablation study.}
We examine the impact of \model{} components by comparing the performance of its variations.
The results of \model{} (\textit{w/o} $Neg_{STMS}$) suggest that the utilization of noisy negative data from STMS resulted in a significant improvement in precision.
At the same time, we observed that \model{} (\textit{w/o} $Neg_{STMS}$) achieved fairly high recall on both datasets.
This can be attributed to the unknown drug jargon in STMS, being labeled as negatives due to being outside of the seed terms during the distant supervision setting.
The effect of pretraining was significant, as \model{} shows performance improvements in both precision and recall by pretraining.
The enhancement in recall was especially notable, suggesting that domain knowledge obtained from pretraining on drug discussions aided in detecting more challenging and diverse drug jargon.
When comparing \model{} and \model{} (\textit{w/o word}), we find that utilizing word information led to improvements in precision.
% \diff{However, when we compare the two models without pretraining (\model{} (\textit{w/o pretrain}) and \model{} (\textit{w/o pretrain\&word})), we find that the inclusion of the word information leads to negligible impact on F1-score despite a slight increase in precision.}
However, when comparing two non-pretrained models (\model{} (\textit{w/o pretrain}) and \model{} (\textit{w/o pretrain\&word})), using word information leads to a decrease in recall and only a slight improvement in precision.
% variation to examine the effect of the word information without using pretraining, we observed a slight increase in precision across both datasets when utilizing the word information. However, this was accompanied by a decrease in recall and a consequent decline in the overall F1 score.
This suggests that when utilizing word embeddings from an off-the-shelf BERT model, which lacks adequate knowledge of drug contexts, the embeddings can actually hinder the detection of a diverse range of drug jargon.

From the results, we suggest the effectiveness of context-based approaches in content moderation.
We show \model{} has superior performance on the jargon detection task compared to other baselines.
% We also show \model{} has superior performance in jargon detection compared to other baselines, and explore the impact of information derived from diverse settings of \model{}.
% We also investigate the input features that can impact \model{}'s performance and suggest how they should be selected for practical application in Appendix~\ref{further_exps}.
% \diff{We also demonstrate the effectiveness of \model{} in a cross-domain setting and the effect of cross-validation in \model{} training in Appendix~\ref{cross_domain} and ~\ref{cross_validation}, respectively.}
We also demonstrate the effectiveness of \model{} in a cross-domain setting in Appendix~\ref{cross_domain}.

\subsection{Qualitative Analysis}
\label{qaulitative_evaluation}
We further conduct qualitative analysis to investigate the limitations of each moderation approach.
In this task, we compare the generated lists of detected drug jargon by each approach. We compare \model{} and the four word-based methods trained in Section~\ref{experimental_setup}.

\vspace{0.1cm}
\noindent\textbf{Drug jargon list construction.}
We construct a list of words by applying \model{} to all candidates (alphanumeric nouns) in the \textit{entire corpus}.
% ~\footnote{To evaluate the efficiency of each approach across a broader range of contexts, we conducted qualitative analysis by applying \model{} to the entire corpus, not limited to the evaluation datasets.}
To score a word candidate, we count both prediction frequency $F_{pred}$ (\# positive predictions) and prediction ratio $R_{pred}$ (\# positive predictions divided by \# total predictions made) of the word. Our score is given by the following formula: 
$\textit{Score} = F_{pred}*{R_{pred}}^{n}$.
% \begin{align}
%     \textit{Score} = F_{pred}*{R_{pred}}^{n}
% \end{align}
The weighing parameter $n$ adjusts sensitivity to frequently occurring words. We set $n = 2$ and extracted the top 100 words. 
For word-based methods, we also select the top 100 words.

\vspace{0.1cm}
\noindent\textbf{Analysis.}
For each word, we manually investigate if the word was used as drug jargon in the corpus (using the guidelines established before). 
We present the top 100 extracted jargon list of each approach in our \href{\git}{repository}.
We also identify cases of failure of the baseline approaches and present examples in Table~\ref{tab:qaulitative_examples}.

The Word2Vec methods showed high performance in precision.
However, Word2Vec may not be able to effectively identify euphemisms since it assumes a single meaning for each word.
We find that the Word2Vec list missed many euphemisms found by other approaches.
For instance, Word2Vec could not identify \textit{weed}, \textit{speed}, \textit{lucy}, \textit{spice}, and \textit{crack}, which were all identified by both \model{} and Zhu et al.'s model.
Therefore, adversaries can easily exploit Word2Vec's weakness by using common words as euphemisms.

CantReader showed low overall precision with many false positives.
CantReader is designed to find euphemisms by finding words that are used differently across corpora. However, with the exception of \textit{weed}, it was also not able to find the above euphemisms that Word2Vec failed to find.
The CantReader word list reveals that many false positives came from misspelled non-jargon words, such as \textit{superawesome}, \textit{comfused}, and \textit{niceeeeee}.
By design, the system gives infrequent words high scores since they may be characteristic to the target corpus.
However, this suggests it is unsuitable to use with user-generated texts, which often contain typos.

Zhu et al.'s approach is based on making a pretrained BERT model that predicts how likely each word is able to fit a context.
Although this model is capable of finding many euphemistic terms, words outside of BERT's limited wordpiece vocabulary cannot be detected by this approach. 
Terms like \textit{valium}, \textit{ambien}, \textit{diazepam}, \textit{oxy}, and \textit{focalin} were terms found by other methods, but were not found by Zhu et al.'s model because they are not in BERT's vocabulary.
This limitation also causes false positives.
BERT splits out-of-vocabulary words into wordpieces, so unknown words (like \textit{meth}) will be represented as a combination of two words (\textit{met} and \textit{\#\#h}).
The false positives \textit{met}, \textit{add}, and \textit{mx} may indicate \textit{meth}, \textit{adderall}, and \textit{mxe}.
This highlights the limitation that the model cannot predict out-of-vocabulary words correctly, even when the model has sufficiently learned on them.

\model{} judges words mainly by how their contexts appear to be. Therefore, a common type of error was from legal drugs that are often discussed in similar ways to illicit drugs, such as \textit{alcohol} and \textit{caffeine}.
However, these errors were common in other approaches as well, suggesting it is a general linguistic limitation, which we elaborate on in Section~\ref{challenges}.
\model{}, unlike other baselines, had no significant systematic failures in recall.
Qualitative analysis suggests \model{} exhibits resilience to a variety of evasion strategies, while maintaining extensive detection coverage.
\section{Discussion}
\label{discussion}
\subsection{Linguistic Challenges}
\label{challenges}
We have argued for a delexicalized approach to automate content moderation.
However, an entirely delexicalized detection has its limitations.
Previous work noted that not all contexts of drug jargon are indicative of drug activity~\cite{zhu2021self}.
For instance, the sentence ``Oh boy do I love ketamine'' from our corpus features the drug \textit{ketamine} in a non-indicative context.
A challenge for a detection system is to understand that the context ``Oh boy do I love [MASK]'' can potentially accommodate both drug jargon (\textit{ketamine}) and non-drug words (\textit{Fridays}).
For these ambiguous cases, lexical information of the masked word is required for an accurate prediction.
\model{} successfully addresses these cases by ensembling delexicalized prediction with lexically informed prediction.

Another challenge is to make the distinction between proper and improper usages of legal drugs.
Legal substances such as \textit{caffeine} or \textit{alcohol} may be consumed irresponsibly (``I'm already addicted to \textit{alcohol}''), and can cause false positives.
While these false positives can be filtered relatively easily with an ``allowlist'', the problem extends to prescription drugs.
Prescription drugs such as \textit{vyvanse} have both legal and abusive usages.
Some contexts may explicitly imply legal usage (``I've been prescribed Vyvanse for the past 2 years''), but many discussions on these drugs (``Will \textit{Vyvanse} sabotage the [workout] progress I'm making'') may not contain enough information to evaluate the legality of usage.
It could be a further challenge to identify sentences of ambiguous legality.
Whether to permit or moderate these cases could be a matter of policy. 

% \begin{figure}[t]
%     \centering
%     \includegraphics[width=0.99\linewidth]{Figures/Application_scenario.pdf}
%     \caption{\model{} workflow in the application scenario.}
%     \label{fig:application}
% \end{figure}

\subsection{Effects of \model{} Configurations}
\label{further_exps}
% Figure~\ref{fig:application} shows the workflow of \model{} in the application scenario. 
Effective moderation using \model{} requires careful configurations. Thus, we analyze various configuration aspects and find insights into how they affect the overall performance.
We also discuss the potential application of \model{} to other fields in Appendix~\ref{other_fields}.

\begin{table}[t]
    \centering
    % \caption{Jargon detection results on Reddit Drug evaluation dataset with modified train settings. $\model{}_{unamb}$ and $\model{}_{control}$ represent models trained on only unambiguous data and the randomly sampled control data, respectively.}
    \caption{\diff{Jargon detection results on the Reddit Drug dataset. $\model{}_{unamb}$ is trained on unambiguous data, while $\model{}_{control}$ uses randomly sampled control data.}}
    \footnotesize
    % \resizebox{0.97\linewidth}{!}{%
        \begin{tabular} {l c c c c}
            \toprule
             & \textbf{\# Positive data} &\textbf{Precision}  & \textbf{Recall}   & \textbf{F1-score} \\
            \midrule
            \vspace{0.05cm}
            \textbf{$\model{}_{original}$} & {329,573} & \diff{0.6659} &  \diff{0.6197} &  \diff{0.6419} \\
            \vspace{0.05cm}
            \textbf{$\model{}_{unamb}$} & {46,325} & \diff{0.7949} & \diff{0.2774} & \diff{0.4113} \\
            % \vspace{0.05cm}
            \textbf{$\model{}_{control}$} & {46,325} & \diff{0.5966} & \diff{0.6219} & \diff{0.6090} \\
            \bottomrule
        \end{tabular}%
    % }
    \vspace{-0.1cm}
    \label{tab:evaluation_informative}
\end{table}

% \begin{table}[t]
%     \centering
%     % \caption{Jargon detection results on Reddit Drug evaluation dataset with modified train settings. $\model{}_{unamb}$ and $\model{}_{control}$ represent models trained on only unambiguous data and the randomly sampled control data, respectively.}
%     \caption{\diff{Jargon detection results on the Reddit Drug dataset. $\model{}_{unamb}$ is trained on unambiguous data, while $\model{}_{control}$ uses randomly sampled control data.}}
%     \footnotesize
%     % \resizebox{0.97\linewidth}{!}{%
%         \begin{tabular} {l c c c c}
%             \toprule
%              & \textbf{\# Positive data} &\textbf{Precision}  & \textbf{Recall}   & \textbf{F1-score} \\
%             \midrule
%             \vspace{0.05cm}
%             \textbf{$\model{}_{original}$} & 329,573 & 0.6659 &  0.6197 &  0.6419 \\
%             \vspace{0.05cm}
%             \textbf{$\model{}_{unamb}$} & 46,325 & 0.7949 & 0.2774 & 0.4113 \\
%             % \vspace{0.05cm}
%             \textbf{$\model{}_{control}$} & 46,325 & 0.5966 & 0.6219 & 0.6090 \\
%             \bottomrule
%         \end{tabular}%
%     % }
%     \label{tab:evaluation_informative}
% \end{table}
\vspace{0.1cm}
\noindent\textbf{Effect of ambiguous contexts.}
As discussed in Section~\ref{challenges}, drug jargon can emerge in ambiguous contexts, which might complicate their usage as training data by introducing confusing samples. Addressing this, Zhu et al.~\cite{zhu2021self} suggested a filtering technique for such contexts and demonstrated enhanced performance when leveraging only unambiguous contexts for training. 
To examine this effect on \model{}, we adopt Zhu et al.'s technique, trimming our potentially ambiguous training data to only unambiguous samples. 
As a control, we also create a dataset of equivalent size from the unfiltered data. With the same hyperparameters for negative training data, we assess jargon detection performance on these new training datasets, as detailed in Table~\ref{tab:evaluation_informative}.
The model trained on only unambiguous data suffered a huge drop in overall performance, while the control model only marginally dropped.
When trained without ambiguous data, the model was able to achieve improvements in precision, but drastically lost recall.
The results suggest that the model trained only on clear-cut positive data cannot generalize fully to the diverse contexts that drug jargon appears.
We hypothesize that a model that understands the potential illicitness of ambiguous contexts is generally better than one that disregards them completely.

\begin{figure}[t]
    \centering
    \begin{subfigure}[t]{0.45\linewidth}
        \centering
        \includegraphics[width=\linewidth]{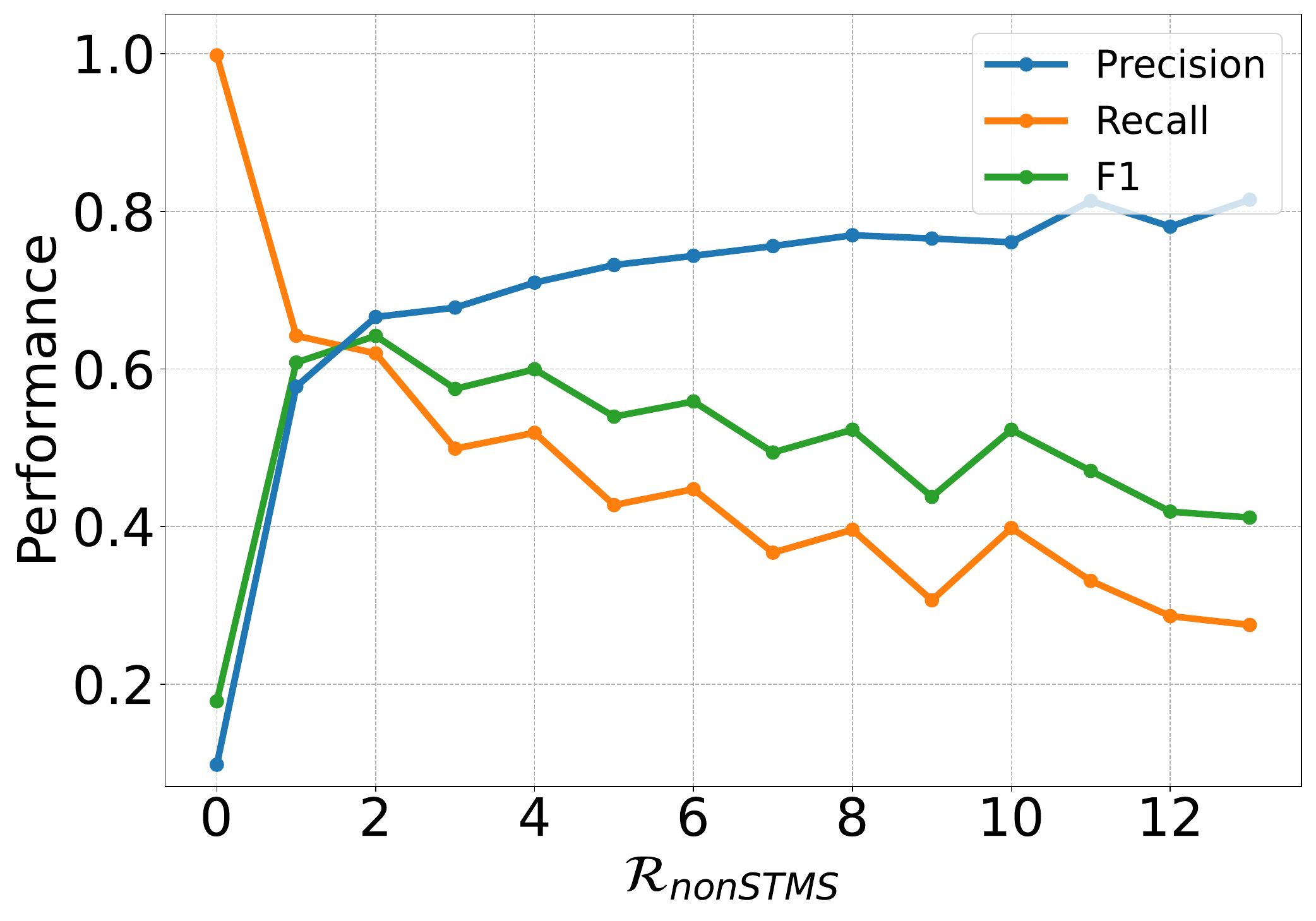}
        \caption{Reddit Drug}
        \label{fig:reddit_perf_on_easy_neg}
    \end{subfigure}
    \hspace{3mm}
    \begin{subfigure}[t]{0.45\linewidth}
        \centering
        \includegraphics[width=\linewidth]{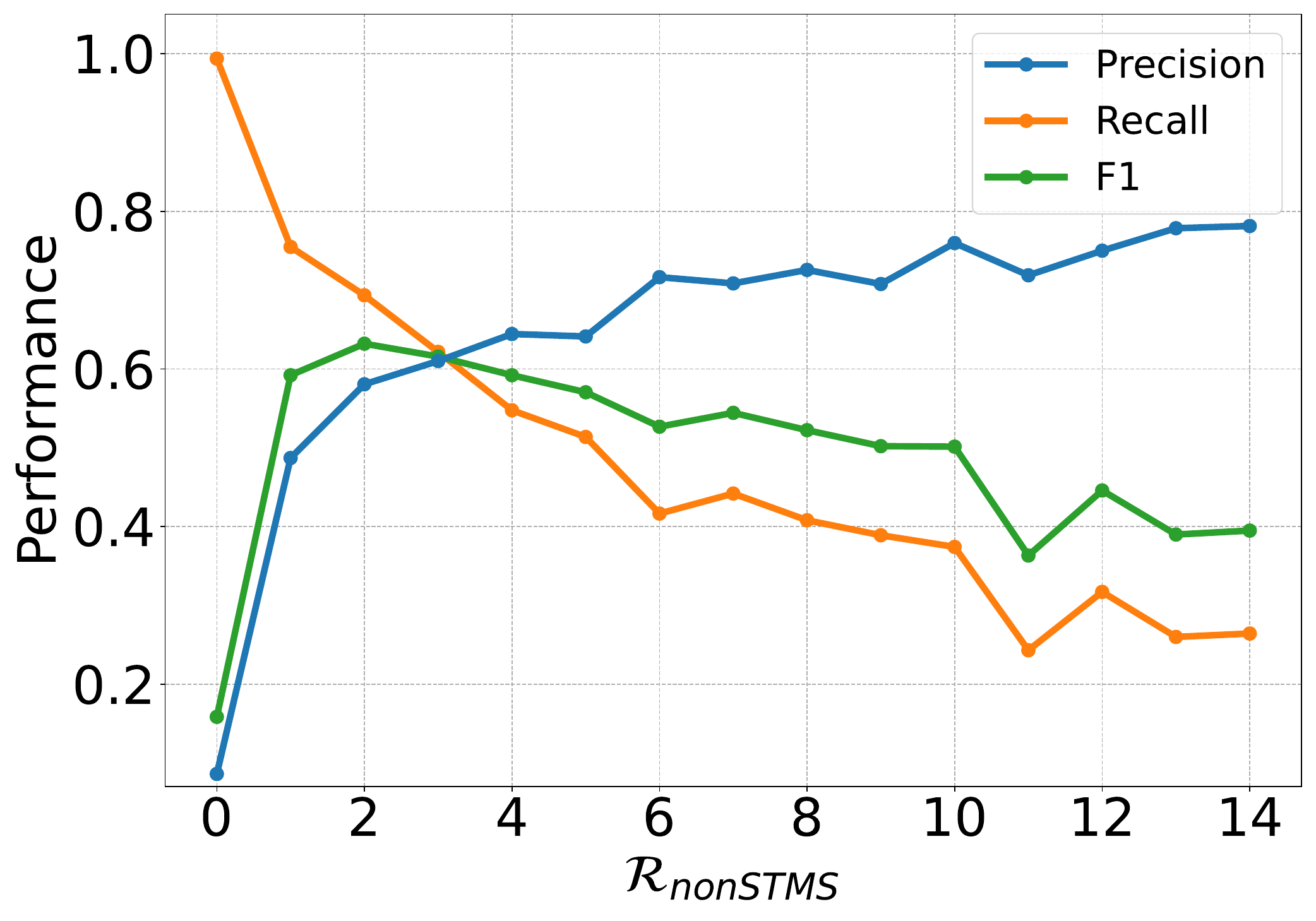}
        \caption{Silk Road Forum}
        \label{fig:sr1_perf_on_easy_neg}
    \end{subfigure}
    \caption{\diff{Jargon detection performance with varying values of $\mathcal{R}_{nonSTMS}$. $\mathcal{R}_{STMS}$ is set to 5.}}
    \label{fig:perf_on_easy_neg}
    \vspace{-0.1cm}
\end{figure}

\vspace{0.1cm}
\noindent\textbf{Effect of negative ratios.}
% Distant supervision is an effective training method, even if it produces potentially noisy labels.
If we rely too heavily on distant supervision (i.e., label every other context as negatives), jargon outside the seed term list will be falsely marked as negatives.
Therefore, we select only a portion of the available context to use as negatives by choosing a hyperparameter $\mathcal{R}_{nonSTMS}$.
Figure~\ref{fig:perf_on_easy_neg} illustrates \model{}'s performance with varying $\mathcal{R}_{nonSTMS}$ values, which can control the influence of potentially noisy negative labels. 
Models trained without $Neg_{nonSTMS}$ exhibit low precision and high recall, indicating a bias towards positive predictions. While negative training sentences are essential, an excess can degrade performance, likely from noisy label introduction. A rise in $\mathcal{R}_{nonSTMS}$ boosts precision but diminishes recall, indicating a tradeoff determined by the volume of negative training data.
For both datasets, having $\mathcal{R}_{nonSTMS}$ = 2 produced the best F1-score.
However, $\mathcal{R}_{nonSTMS}$ can further be tuned depending on whether a model is designed to be \textit{precision-sensitive} or \textit{recall-sensitive}.

\begin{figure}[t]
    \centering
    \begin{subfigure}[t]{0.45\linewidth}
        \centering
        \includegraphics[width=\linewidth]{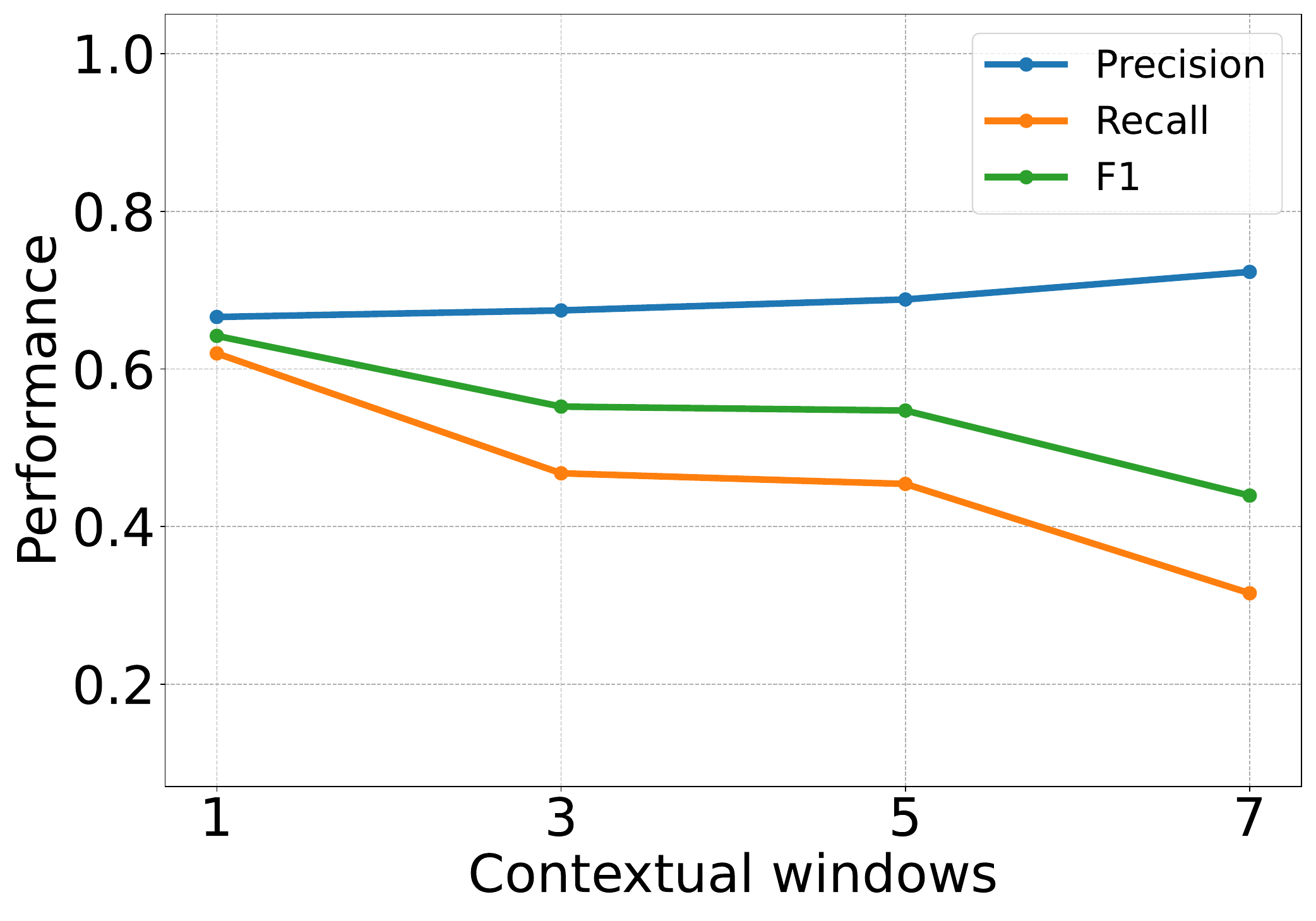}
        \caption{Reddit Drug}
        \label{fig:reddit_perf_on_contextual}
    \end{subfigure}
    \hspace{3mm}
    \begin{subfigure}[t]{0.45\linewidth}
        \centering
        \includegraphics[width=\linewidth]{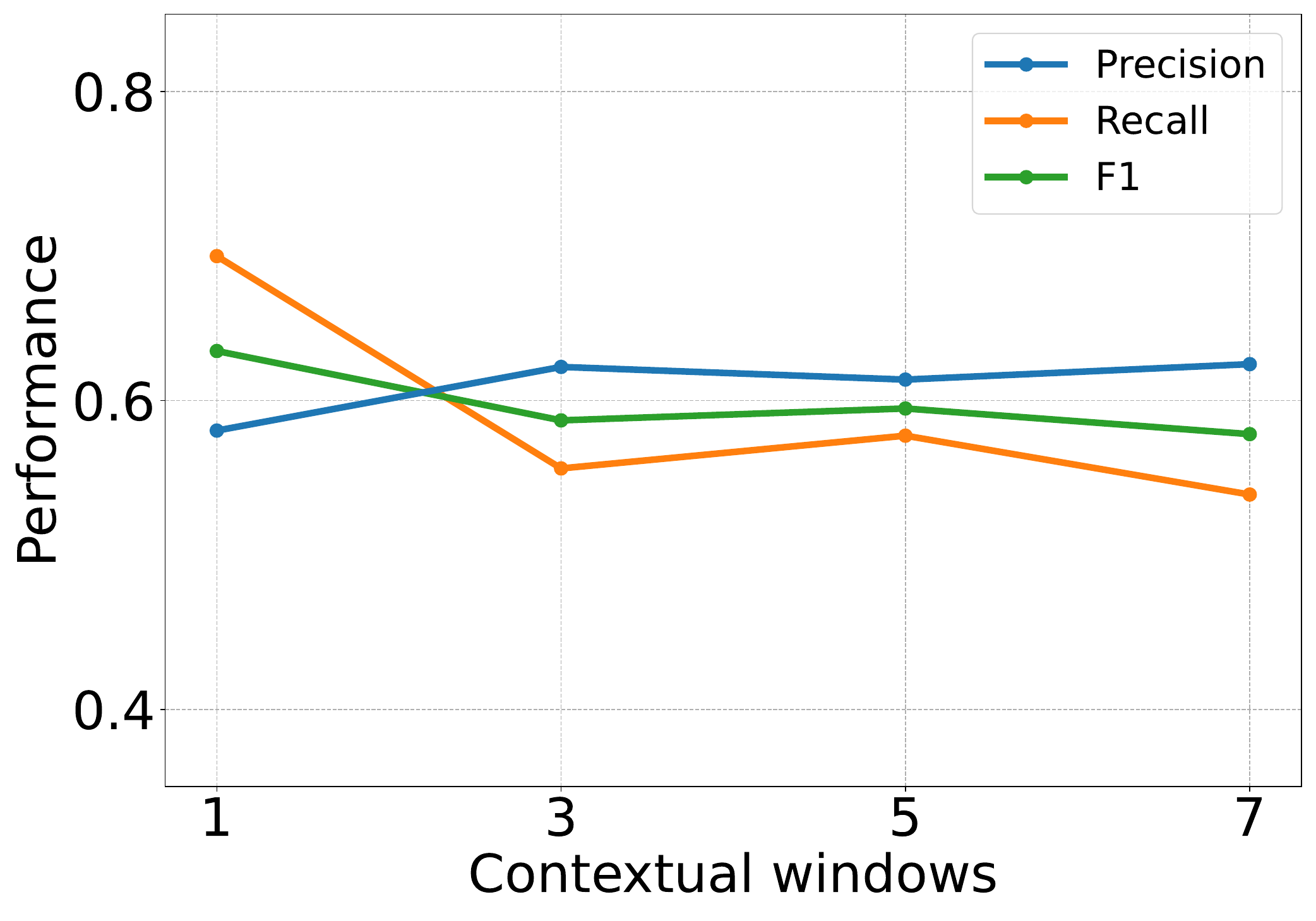}
        \caption{Silk Road Forum}
        \label{fig:sr1_perf_on_contextual}
    \end{subfigure}
    \caption{\diff{Jargon detection performance with varying contextual window sizes.}}
    \label{fig:perf_on_contextual_window}
    \vspace{-0.1cm}
\end{figure}
\vspace{0.1cm}
\diff{
\noindent\textbf{Effect of contextual windows.}
\model{} conducts detection at the single-sentence level.
To explore the impact of broader contexts, we expanded the contextual window to include up to three additional sentences on either side of the target sentence, increasing the window sizes to encompass 3, 5, and 7 sentences in total.
As shown in Figure~\ref{fig:perf_on_contextual_window}, this expansion improved precision by providing more contextual clues for more accurate detection.
However, we observed a significant reduction in recall and F1-score, primarily due to the amplification of false negatives. 
This issue arises from the use of noisy negatives through distant supervision, which becomes more pronounced with larger windows.
Moreover, larger windows increased computational overhead during training and inference. 
Given these factors, we concluded a single-sentence window is the optimal approach in configuring \model{}.
}

\vspace{0.1cm}
\noindent\textbf{Effect of seed terms and iterative use of \model{}.}
\begin{table}[t]
    \centering
    \caption{Jargon detection results on Reddit Drug evaluation dataset with varying numbers of seed terms.}
    \footnotesize
    % \resizebox{0.99\linewidth}{!}{%
        \begin{tabular} {l c  c c}
            \toprule
            \textbf{$\model{}_{\text{\# seed terms}}$} &\textbf{Precision}  & \textbf{Recall}   & \textbf{F1-score} \\
            \midrule
            \vspace{0.05cm}
            \textbf{$\model{}_{23}$} & 0.6039 & 0.3445 & 0.4387 \\
            \vspace{0.05cm}
            \textbf{$\model{}_{46}$} & 0.6659 & 0.6197 &  0.6419 \\
            % \vspace{0.05cm}
            \textbf{$\model{}_{70}$} & 0.6781 & 0.6242 & 0.6503 \\
            \bottomrule
        \end{tabular}%
    % }
    \label{tab:evaluation_seedterms}
\end{table}

We examine the effect of drug seed terms on \model{} by repeating experiments with reduced and extended seed term lists.
First, we train a model with a reduced seed term list of 23 words (randomly selected from the original list of 46 terms).
Then, we train another model with an extended 64-word seed term list, which was created by adding 18 new words that \model{} correctly found.
The results are shown in Table~\ref{tab:evaluation_seedterms}. 
\model{} trained on the reduced list achieves a worse result than the original, especially in recall.
This suggests \model{} is indeed sensitive to the seed terms used to create distant supervision.
However, the model trained on the reduced list still manages to reach a reasonable performance.
It could be used to bootstrap a larger list of seed terms by discovering new jargon.
Although human effort is required to identify quality terms, this can facilitate the development of a seed list comparable to the original, achieving similar performance. 
Similarly, the original model can be used to iteratively increment the seed term list with newly detected words. When \model{} was trained on the extended list, we observe that the performance increased by a relatively small amount. This suggests \model{} can be improved with a larger list of seed terms, but with diminishing returns.
This iterative updating process illustrates \model{}'s capacity to adapt to evolving landscapes, continuously refining its performance by integrating new seed terms.
% \balance
\vspace{0.27cm}
\section{Conclusion}
Automatic content moderation of drugs on social media platforms is an important but challenging task.
We present the drug jargon detection framework \model{}, which can handle unseen terms and euphemisms more effectively than previously proposed methods.
\model{} achieves this through a novel delexicalized distant supervision training method on an unannotated corpus.
Our evaluations on two manually annotated datasets show that \model{} outperforms state-of-the-art baselines.
Through qualitative analysis, we show that \model{} also has superior coverage and robustness.

\section*{Acknowledgement}
This work was supported by Institute of Information \& communications Technology Planning \& Evaluation (IITP) grant funded by the Korea government (MSIT) (RS- 2023-00215700, Trustworthy Metaverse: blockchain-enabled convergence research, 50\%), and S2W Inc. through financial support and technical assistance (50\%).

\bibliographystyle{ACM-Reference-Format}
\bibliography{references}


\begin{thebibliography}{54}

%%% the Web address.

\ifx \showCODEN    \undefined \def \showCODEN     #1{\unskip}     \fi
\ifx \showDOI      \undefined \def \showDOI       #1{#1}\fi
\ifx \showISBNx    \undefined \def \showISBNx     #1{\unskip}     \fi
\ifx \showISBNxiii \undefined \def \showISBNxiii  #1{\unskip}     \fi
\ifx \showISSN     \undefined \def \showISSN      #1{\unskip}     \fi
\ifx \showLCCN     \undefined \def \showLCCN      #1{\unskip}     \fi
\ifx \shownote     \undefined \def \shownote      #1{#1}          \fi
\ifx \showarticletitle \undefined \def \showarticletitle #1{#1}   \fi
\ifx \showURL      \undefined \def \showURL       {\relax}        \fi
% invisible to TeX
\providecommand\bibfield[2]{#2}
\providecommand\bibinfo[2]{#2}
\providecommand\natexlab[1]{#1}
\providecommand\showeprint[2][]{arXiv:#2}

\bibitem[Administration(2018)]%
        {dea_list}
\bibfield{author}{\bibinfo{person}{Drug~Enforcement Administration}.} \bibinfo{year}{2018}\natexlab{}.
\newblock \bibinfo{title}{{Slang Terms and Code Words: A Reference for Law Enforcement Personnel}}.
\newblock \bibinfo{howpublished}{\url{https://www.dea.gov/sites/default/files/2018-07/DIR-022-18.pdf}}.
\newblock
\newblock
\shownote{Accessed: 2022-10-26}.


\bibitem[Branwen et~al\mbox{.}(2015)]%
        {dnmArchives}
\bibfield{author}{\bibinfo{person}{Gwern Branwen}, \bibinfo{person}{Nicolas Christin}, \bibinfo{person}{David Décary-Hétu}, \bibinfo{person}{Rasmus~Munksgaard Andersen}, \bibinfo{person}{StExo}, \bibinfo{person}{El Presidente}, \bibinfo{person}{Anonymous}, \bibinfo{person}{Daryl Lau}, \bibinfo{person}{Delyan~Kratunov Sohhlz}, \bibinfo{person}{Vince Cakic}, \bibinfo{person}{Van Buskirk}, \bibinfo{person}{Whom}, \bibinfo{person}{Michael McKenna}, {and} \bibinfo{person}{Sigi Goode}.} \bibinfo{year}{2015}\natexlab{}.
\newblock \bibinfo{title}{{Dark Net Market archives, 2011-2015}}.
\newblock \bibinfo{howpublished}{\url{https://www.gwern.net/DNM-archives}}.
\newblock
\urldef\tempurl%
\url{https://www.gwern.net/DNM-archives}
\showURL{%
\tempurl}
\newblock
\shownote{Accessed: 2022-10-26}.


\bibitem[Cone(2006)]%
        {drug_bias}
\bibfield{author}{\bibinfo{person}{Edward~J. Cone}.} \bibinfo{year}{2006}\natexlab{}.
\newblock \showarticletitle{Ephemeral profiles of prescription drug and formulation tampering: Evolving pseudoscience on the Internet}.
\newblock \bibinfo{journal}{\emph{Drug and Alcohol Dependence}}  \bibinfo{volume}{83} (\bibinfo{year}{2006}), \bibinfo{pages}{S31--S39}.
\newblock
\showISSN{0376-8716}
\urldef\tempurl%
\url{https://doi.org/10.1016/j.drugalcdep.2005.11.027}
\showDOI{\tempurl}
\newblock
\shownote{Drug Formulation and Abuse Liability}.


\bibitem[Devlin et~al\mbox{.}(2018)]%
        {devlin2018bert}
\bibfield{author}{\bibinfo{person}{Jacob Devlin}, \bibinfo{person}{Ming-Wei Chang}, \bibinfo{person}{Kenton Lee}, {and} \bibinfo{person}{Kristina Toutanova}.} \bibinfo{year}{2018}\natexlab{}.
\newblock \showarticletitle{Bert: Pre-training of deep bidirectional transformers for language understanding}.
\newblock \bibinfo{journal}{\emph{arXiv preprint arXiv:1810.04805}} (\bibinfo{year}{2018}).
\newblock


\bibitem[for Disease~Control and Prevention(2022)]%
        {od_deaths2}
\bibfield{author}{\bibinfo{person}{Centers for Disease~Control} {and} \bibinfo{person}{Prevention}.} \bibinfo{year}{2022}\natexlab{}.
\newblock \bibinfo{title}{{Drug Overdose Deaths in the United States, 2001–2021}}.
\newblock \bibinfo{howpublished}{\url{https://www.cdc.gov/nchs/products/databriefs/db457.htm}}.
\newblock
\newblock
\shownote{Accessed: 2023-02-07}.


\bibitem[Franceschi-Bicchierai(2018)]%
        {reddit_ban}
\bibfield{author}{\bibinfo{person}{Lorenzo Franceschi-Bicchierai}.} \bibinfo{year}{2018}\natexlab{}.
\newblock \bibinfo{title}{{Reddit Bans Subreddits Dedicated to Dark Web Drug Markets and Selling Guns}}.
\newblock \bibinfo{howpublished}{\url{https://www.vice.com/en/article/ne9v5k/reddit-bans-subreddits-dark-web-drug-markets-and-guns}}.
\newblock
\newblock
\shownote{Accessed: 2022-10-26}.


\bibitem[Gururangan et~al\mbox{.}(2020)]%
        {dapt}
\bibfield{author}{\bibinfo{person}{Suchin Gururangan}, \bibinfo{person}{Ana Marasovi{\'c}}, \bibinfo{person}{Swabha Swayamdipta}, \bibinfo{person}{Kyle Lo}, \bibinfo{person}{Iz Beltagy}, \bibinfo{person}{Doug Downey}, {and} \bibinfo{person}{Noah~A. Smith}.} \bibinfo{year}{2020}\natexlab{}.
\newblock \showarticletitle{Don{'}t Stop Pretraining: Adapt Language Models to Domains and Tasks}. In \bibinfo{booktitle}{\emph{Proceedings of the 58th Annual Meeting of the Association for Computational Linguistics}}. \bibinfo{publisher}{Association for Computational Linguistics}, \bibinfo{address}{Online}, \bibinfo{pages}{8342--8360}.
\newblock
\urldef\tempurl%
\url{https://doi.org/10.18653/v1/2020.acl-main.740}
\showDOI{\tempurl}


\bibitem[Hine et~al\mbox{.}(2017)]%
        {keks}
\bibfield{author}{\bibinfo{person}{Gabriel~Emile Hine}, \bibinfo{person}{Jeremiah Onaolapo}, \bibinfo{person}{Emiliano~De Cristofaro}, \bibinfo{person}{Nicolas Kourtellis}, \bibinfo{person}{Ilias Leontiadis}, \bibinfo{person}{Riginos Samaras}, \bibinfo{person}{Gianluca Stringhini}, {and} \bibinfo{person}{Jeremy Blackburn}.} \bibinfo{year}{2017}\natexlab{}.
\newblock \showarticletitle{Kek, Cucks, and God Emperor Trump: A Measurement Study of 4chan's Politically Incorrect Forum and Its Effects on the Web}. In \bibinfo{booktitle}{\emph{ICWSM}}.
\newblock


\bibitem[Hoffa(2015)]%
        {reddit_query}
\bibfield{author}{\bibinfo{person}{Felipe Hoffa}.} \bibinfo{year}{2015}\natexlab{}.
\newblock \bibinfo{title}{{1.7 billion reddit comments loaded on BigQuery}}.
\newblock \bibinfo{howpublished}{\url{https://www.reddit.com/r/bigquery/comments/3cej2b/17_billion_reddit_comments_loaded_on_bigquery/}}.
\newblock
\newblock
\shownote{Accessed: 2022-10-25}.


\bibitem[Hu et~al\mbox{.}(2021)]%
        {insta_multimodal}
\bibfield{author}{\bibinfo{person}{Chuanbo Hu}, \bibinfo{person}{Minglei Yin}, \bibinfo{person}{Bin Liu}, \bibinfo{person}{Xin Li}, {and} \bibinfo{person}{Yanfang Ye}.} \bibinfo{year}{2021}\natexlab{}.
\newblock \showarticletitle{Detection of Illicit Drug Trafficking Events on Instagram: A Deep Multimodal Multilabel Learning Approach}. In \bibinfo{booktitle}{\emph{Proceedings of the 30th ACM International Conference on Information \& Knowledge Management}}. \bibinfo{pages}{3838--3846}.
\newblock


\bibitem[Jin et~al\mbox{.}(2023)]%
        {jin2023darkbert}
\bibfield{author}{\bibinfo{person}{Youngjin Jin}, \bibinfo{person}{Eugene Jang}, \bibinfo{person}{Jian Cui}, \bibinfo{person}{Jin-Woo Chung}, \bibinfo{person}{Yongjae Lee}, {and} \bibinfo{person}{Seungwon Shin}.} \bibinfo{year}{2023}\natexlab{}.
\newblock \showarticletitle{DarkBERT: A Language Model for the Dark Side of the Internet}. In \bibinfo{booktitle}{\emph{Proceedings of the 61st Annual Meeting of the Association for Computational Linguistics (Volume 1: Long Papers)}}. \bibinfo{pages}{7515--7533}.
\newblock


\bibitem[Jin et~al\mbox{.}(2022)]%
        {coda}
\bibfield{author}{\bibinfo{person}{Youngjin Jin}, \bibinfo{person}{Eugene Jang}, \bibinfo{person}{Yongjae Lee}, \bibinfo{person}{Seungwon Shin}, {and} \bibinfo{person}{Jin-Woo Chung}.} \bibinfo{year}{2022}\natexlab{}.
\newblock \showarticletitle{Shedding New Light on the Language of the Dark Web}. In \bibinfo{booktitle}{\emph{Proceedings of the 2022 Conference of the North American Chapter of the Association for Computational Linguistics: Human Language Technologies}}. \bibinfo{publisher}{Association for Computational Linguistics}, \bibinfo{address}{Seattle, United States}, \bibinfo{pages}{5621--5637}.
\newblock
\urldef\tempurl%
\url{https://doi.org/10.18653/v1/2022.naacl-main.412}
\showDOI{\tempurl}


\bibitem[Ke et~al\mbox{.}(2022)]%
        {chinese_mlm}
\bibfield{author}{\bibinfo{person}{Liang Ke}, \bibinfo{person}{Xinyu Chen}, {and} \bibinfo{person}{Haizhou Wang}.} \bibinfo{year}{2022}\natexlab{}.
\newblock \showarticletitle{An Unsupervised Detection Framework for Chinese Jargons in the Darknet}. In \bibinfo{booktitle}{\emph{Proceedings of the Fifteenth ACM International Conference on Web Search and Data Mining}} (Virtual Event, AZ, USA) \emph{(\bibinfo{series}{WSDM '22})}. \bibinfo{publisher}{Association for Computing Machinery}, \bibinfo{address}{New York, NY, USA}, \bibinfo{pages}{458–466}.
\newblock
\showISBNx{9781450391320}
\urldef\tempurl%
\url{https://doi.org/10.1145/3488560.3498469}
\showDOI{\tempurl}


\bibitem[Kirk et~al\mbox{.}(2022)]%
        {kirk-etal-2022-hatemoji}
\bibfield{author}{\bibinfo{person}{Hannah Kirk}, \bibinfo{person}{Bertie Vidgen}, \bibinfo{person}{Paul Rottger}, \bibinfo{person}{Tristan Thrush}, {and} \bibinfo{person}{Scott Hale}.} \bibinfo{year}{2022}\natexlab{}.
\newblock \showarticletitle{{H}atemoji: A Test Suite and Adversarially-Generated Dataset for Benchmarking and Detecting Emoji-Based Hate}. In \bibinfo{booktitle}{\emph{Proceedings of the 2022 Conference of the North American Chapter of the Association for Computational Linguistics: Human Language Technologies}}. \bibinfo{publisher}{Association for Computational Linguistics}, \bibinfo{address}{Seattle, United States}, \bibinfo{pages}{1352--1368}.
\newblock
\urldef\tempurl%
\url{https://doi.org/10.18653/v1/2022.naacl-main.97}
\showDOI{\tempurl}


\bibitem[Lee et~al\mbox{.}(2022)]%
        {lee-etal-2022-searching}
\bibfield{author}{\bibinfo{person}{Patrick Lee}, \bibinfo{person}{Martha Gavidia}, \bibinfo{person}{Anna Feldman}, {and} \bibinfo{person}{Jing Peng}.} \bibinfo{year}{2022}\natexlab{}.
\newblock \showarticletitle{Searching for {PET}s: Using Distributional and Sentiment-Based Methods to Find Potentially Euphemistic Terms}. In \bibinfo{booktitle}{\emph{Proceedings of the Second Workshop on Understanding Implicit and Underspecified Language}}, \bibfield{editor}{\bibinfo{person}{Valentina Pyatkin}, \bibinfo{person}{Daniel Fried}, {and} \bibinfo{person}{Talita Anthonio}} (Eds.). \bibinfo{publisher}{Association for Computational Linguistics}, \bibinfo{address}{Seattle, USA}, \bibinfo{pages}{22--32}.
\newblock
\urldef\tempurl%
\url{https://doi.org/10.18653/v1/2022.unimplicit-1.4}
\showDOI{\tempurl}


\bibitem[Li et~al\mbox{.}(2019)]%
        {insta_text}
\bibfield{author}{\bibinfo{person}{Jiawei Li}, \bibinfo{person}{Qing Xu}, \bibinfo{person}{Neal Shah}, \bibinfo{person}{Tim~K Mackey}, {et~al\mbox{.}}} \bibinfo{year}{2019}\natexlab{}.
\newblock \showarticletitle{A machine learning approach for the detection and characterization of illicit drug dealers on instagram: model evaluation study}.
\newblock \bibinfo{journal}{\emph{Journal of medical Internet research}} \bibinfo{volume}{21}, \bibinfo{number}{6} (\bibinfo{year}{2019}), \bibinfo{pages}{e13803}.
\newblock


\bibitem[Mackey and Kalyanam(2017)]%
        {detect_fent}
\bibfield{author}{\bibinfo{person}{Tim~K Mackey} {and} \bibinfo{person}{Janani Kalyanam}.} \bibinfo{year}{2017}\natexlab{}.
\newblock \showarticletitle{Detection of illicit online sales of fentanyls via Twitter}.
\newblock \bibinfo{journal}{\emph{F1000Research}}  \bibinfo{volume}{6} (\bibinfo{year}{2017}).
\newblock


\bibitem[Mackey et~al\mbox{.}(2017)]%
        {detect_opioid}
\bibfield{author}{\bibinfo{person}{Tim~K Mackey}, \bibinfo{person}{Janani Kalyanam}, \bibinfo{person}{Takeo Katsuki}, {and} \bibinfo{person}{Gert Lanckriet}.} \bibinfo{year}{2017}\natexlab{}.
\newblock \showarticletitle{Twitter-based detection of illegal online sale of prescription opioid}.
\newblock \bibinfo{journal}{\emph{American journal of public health}} \bibinfo{volume}{107}, \bibinfo{number}{12} (\bibinfo{year}{2017}), \bibinfo{pages}{1910--1915}.
\newblock


\bibitem[Magu et~al\mbox{.}(2017)]%
        {hatespeech_aaai}
\bibfield{author}{\bibinfo{person}{Rijul Magu}, \bibinfo{person}{Kshitij Joshi}, {and} \bibinfo{person}{Jiebo Luo}.} \bibinfo{year}{2017}\natexlab{}.
\newblock \showarticletitle{Detecting the Hate Code on Social Media}.
\newblock \bibinfo{journal}{\emph{Proceedings of the International AAAI Conference on Web and Social Media}} \bibinfo{volume}{11}, \bibinfo{number}{1} (\bibinfo{date}{May} \bibinfo{year}{2017}), \bibinfo{pages}{608--611}.
\newblock
\urldef\tempurl%
\url{https://doi.org/10.1609/icwsm.v11i1.14921}
\showDOI{\tempurl}


\bibitem[Magu and Luo(2018)]%
        {hatespeech_det}
\bibfield{author}{\bibinfo{person}{Rijul Magu} {and} \bibinfo{person}{Jiebo Luo}.} \bibinfo{year}{2018}\natexlab{}.
\newblock \showarticletitle{Determining Code Words in Euphemistic Hate Speech Using Word Embedding Networks}. In \bibinfo{booktitle}{\emph{Proceedings of the 2nd Workshop on Abusive Language Online ({ALW}2)}}. \bibinfo{publisher}{Association for Computational Linguistics}, \bibinfo{address}{Brussels, Belgium}, \bibinfo{pages}{93--100}.
\newblock
\urldef\tempurl%
\url{https://doi.org/10.18653/v1/W18-5112}
\showDOI{\tempurl}


\bibitem[Manolache et~al\mbox{.}(2022)]%
        {manolache2022veridark}
\bibfield{author}{\bibinfo{person}{Andrei Manolache}, \bibinfo{person}{Florin Brad}, \bibinfo{person}{Antonio Barbalau}, \bibinfo{person}{Radu~Tudor Ionescu}, {and} \bibinfo{person}{Marius Popescu}.} \bibinfo{year}{2022}\natexlab{}.
\newblock \showarticletitle{VeriDark: A Large-Scale Benchmark for Authorship Verification on the Dark Web}.
\newblock \bibinfo{journal}{\emph{Advances in Neural Information Processing Systems}}  \bibinfo{volume}{35} (\bibinfo{year}{2022}), \bibinfo{pages}{15574--15588}.
\newblock


\bibitem[Meng et~al\mbox{.}(2021)]%
        {distant_ner2}
\bibfield{author}{\bibinfo{person}{Yu Meng}, \bibinfo{person}{Yunyi Zhang}, \bibinfo{person}{Jiaxin Huang}, \bibinfo{person}{Xuan Wang}, \bibinfo{person}{Yu Zhang}, \bibinfo{person}{Heng Ji}, {and} \bibinfo{person}{Jiawei Han}.} \bibinfo{year}{2021}\natexlab{}.
\newblock \showarticletitle{Distantly-supervised named entity recognition with noise-robust learning and language model augmented self-training}.
\newblock \bibinfo{journal}{\emph{arXiv preprint arXiv:2109.05003}} (\bibinfo{year}{2021}).
\newblock


\bibitem[Mikolov et~al\mbox{.}(2013)]%
        {mikolov2013efficient}
\bibfield{author}{\bibinfo{person}{Tomas Mikolov}, \bibinfo{person}{Kai Chen}, \bibinfo{person}{Greg Corrado}, {and} \bibinfo{person}{Jeffrey Dean}.} \bibinfo{year}{2013}\natexlab{}.
\newblock \showarticletitle{Efficient estimation of word representations in vector space}.
\newblock \bibinfo{journal}{\emph{arXiv preprint arXiv:1301.3781}} (\bibinfo{year}{2013}).
\newblock


\bibitem[Min et~al\mbox{.}(2013)]%
        {distant-incomplete-kb}
\bibfield{author}{\bibinfo{person}{Bonan Min}, \bibinfo{person}{Ralph Grishman}, \bibinfo{person}{Li Wan}, \bibinfo{person}{Chang Wang}, {and} \bibinfo{person}{David Gondek}.} \bibinfo{year}{2013}\natexlab{}.
\newblock \showarticletitle{Distant Supervision for Relation Extraction with an Incomplete Knowledge Base}. In \bibinfo{booktitle}{\emph{Proceedings of the 2013 Conference of the North {A}merican Chapter of the Association for Computational Linguistics: Human Language Technologies}}. \bibinfo{publisher}{Association for Computational Linguistics}, \bibinfo{address}{Atlanta, Georgia}, \bibinfo{pages}{777--782}.
\newblock
\urldef\tempurl%
\url{https://aclanthology.org/N13-1095}
\showURL{%
\tempurl}


\bibitem[Mintz et~al\mbox{.}(2009)]%
        {distant_og}
\bibfield{author}{\bibinfo{person}{Mike Mintz}, \bibinfo{person}{Steven Bills}, \bibinfo{person}{Rion Snow}, {and} \bibinfo{person}{Daniel Jurafsky}.} \bibinfo{year}{2009}\natexlab{}.
\newblock \showarticletitle{Distant supervision for relation extraction without labeled data}. In \bibinfo{booktitle}{\emph{Proceedings of the Joint Conference of the 47th Annual Meeting of the {ACL} and the 4th International Joint Conference on Natural Language Processing of the {AFNLP}}}. \bibinfo{publisher}{Association for Computational Linguistics}, \bibinfo{address}{Suntec, Singapore}, \bibinfo{pages}{1003--1011}.
\newblock
\urldef\tempurl%
\url{https://aclanthology.org/P09-1113}
\showURL{%
\tempurl}


\bibitem[Nations({[n.\,d.]})]%
        {united_nations}
\bibfield{author}{\bibinfo{person}{United Nations}.} \bibinfo{year}{[n.\,d.]}\natexlab{}.
\newblock \bibinfo{title}{Break the link between illicit drugs and social media: Un-backed Report | UN News}.
\newblock
\newblock
\urldef\tempurl%
\url{https://news.un.org/en/story/2022/03/1113642}
\showURL{%
\tempurl}


\bibitem[of~Health(2022)]%
        {od_deaths}
\bibfield{author}{\bibinfo{person}{National~Institutes of Health}.} \bibinfo{year}{2022}\natexlab{}.
\newblock \bibinfo{title}{{Overdose Death Rates}}.
\newblock \bibinfo{howpublished}{\url{https://nida.nih.gov/research-topics/trends-statistics/overdose-death-rates}}.
\newblock
\newblock
\shownote{Accessed: 2022-11-08}.


\bibitem[OpenAI(2024)]%
        {gpt4omini}
\bibfield{author}{\bibinfo{person}{OpenAI}.} \bibinfo{year}{2024}\natexlab{}.
\newblock \bibinfo{title}{gpt4omini}.
\newblock \bibinfo{howpublished}{\url{https://platform.openai.com/docs/models/gpt-4o-mini}}.
\newblock
\newblock
\shownote{Accessed: 2024-12-04}.


\bibitem[Pagliery(2013)]%
        {cnn_sr_shutdown}
\bibfield{author}{\bibinfo{person}{Jose Pagliery}.} \bibinfo{year}{2013}\natexlab{}.
\newblock \bibinfo{title}{{FBI shuts down online drug market Silk Road}}.
\newblock \bibinfo{howpublished}{\url{https://money.cnn.com/2013/10/02/technology/silk-road-shut-down/index.html}}.
\newblock
\newblock
\shownote{Accessed: 2022-10-26}.


\bibitem[Qian et~al\mbox{.}(2021)]%
        {insta_meta}
\bibfield{author}{\bibinfo{person}{Yiyue Qian}, \bibinfo{person}{Yiming Zhang}, \bibinfo{person}{Yanfang~(Fa Ye}, {and} \bibinfo{person}{Chuxu Zhang}.} \bibinfo{year}{2021}\natexlab{}.
\newblock \showarticletitle{Distilling Meta Knowledge on Heterogeneous Graph for Illicit Drug Trafficker Detection on Social Media}. In \bibinfo{booktitle}{\emph{Advances in Neural Information Processing Systems}}, \bibfield{editor}{\bibinfo{person}{M.~Ranzato}, \bibinfo{person}{A.~Beygelzimer}, \bibinfo{person}{Y.~Dauphin}, \bibinfo{person}{P.S. Liang}, {and} \bibinfo{person}{J.~Wortman Vaughan}} (Eds.), Vol.~\bibinfo{volume}{34}. \bibinfo{publisher}{Curran Associates, Inc.}, \bibinfo{pages}{26911--26923}.
\newblock
\urldef\tempurl%
\url{https://proceedings.neurips.cc/paper/2021/file/e234e195f3789f05483378c397db1cb5-Paper.pdf}
\showURL{%
\tempurl}


\bibitem[Qin et~al\mbox{.}(2018)]%
        {distant_re}
\bibfield{author}{\bibinfo{person}{Pengda Qin}, \bibinfo{person}{Weiran Xu}, {and} \bibinfo{person}{William~Yang Wang}.} \bibinfo{year}{2018}\natexlab{}.
\newblock \showarticletitle{Robust Distant Supervision Relation Extraction via Deep Reinforcement Learning}. In \bibinfo{booktitle}{\emph{Proceedings of the 56th Annual Meeting of the Association for Computational Linguistics (Volume 1: Long Papers)}}. \bibinfo{publisher}{Association for Computational Linguistics}, \bibinfo{address}{Melbourne, Australia}, \bibinfo{pages}{2137--2147}.
\newblock
\urldef\tempurl%
\url{https://doi.org/10.18653/v1/P18-1199}
\showDOI{\tempurl}


\bibitem[Quirk and Poon(2017)]%
        {quirk-poon-2017-distant}
\bibfield{author}{\bibinfo{person}{Chris Quirk} {and} \bibinfo{person}{Hoifung Poon}.} \bibinfo{year}{2017}\natexlab{}.
\newblock \showarticletitle{Distant Supervision for Relation Extraction beyond the Sentence Boundary}. In \bibinfo{booktitle}{\emph{Proceedings of the 15th Conference of the {E}uropean Chapter of the Association for Computational Linguistics: Volume 1, Long Papers}}. \bibinfo{publisher}{Association for Computational Linguistics}, \bibinfo{address}{Valencia, Spain}, \bibinfo{pages}{1171--1182}.
\newblock
\urldef\tempurl%
\url{https://aclanthology.org/E17-1110}
\showURL{%
\tempurl}


\bibitem[Ray et~al\mbox{.}(2019)]%
        {spoken}
\bibfield{author}{\bibinfo{person}{Avik Ray}, \bibinfo{person}{Yilin Shen}, {and} \bibinfo{person}{Hongxia Jin}.} \bibinfo{year}{2019}\natexlab{}.
\newblock \showarticletitle{Iterative delexicalization for improved spoken language understanding}.
\newblock \bibinfo{journal}{\emph{arXiv preprint arXiv:1910.07060}} (\bibinfo{year}{2019}).
\newblock


\bibitem[Risch et~al\mbox{.}(2020)]%
        {risch-etal-2020-offensive}
\bibfield{author}{\bibinfo{person}{Julian Risch}, \bibinfo{person}{Robin Ruff}, {and} \bibinfo{person}{Ralf Krestel}.} \bibinfo{year}{2020}\natexlab{}.
\newblock \showarticletitle{Offensive Language Detection Explained}. In \bibinfo{booktitle}{\emph{Proceedings of the Second Workshop on Trolling, Aggression and Cyberbullying}}. \bibinfo{publisher}{European Language Resources Association (ELRA)}, \bibinfo{address}{Marseille, France}, \bibinfo{pages}{137--143}.
\newblock
\showISBNx{979-10-95546-56-6}
\urldef\tempurl%
\url{https://aclanthology.org/2020.trac-1.22}
\showURL{%
\tempurl}


\bibitem[Rönkä and Katainen(2017)]%
        {nonmedical}
\bibfield{author}{\bibinfo{person}{Sanna Rönkä} {and} \bibinfo{person}{Anu Katainen}.} \bibinfo{year}{2017}\natexlab{}.
\newblock \showarticletitle{Non-medical use of prescription drugs among illicit drug users: A case study on an online drug forum}.
\newblock \bibinfo{journal}{\emph{International Journal of Drug Policy}}  \bibinfo{volume}{39} (\bibinfo{year}{2017}), \bibinfo{pages}{62--68}.
\newblock
\showISSN{0955-3959}
\urldef\tempurl%
\url{https://doi.org/10.1016/j.drugpo.2016.08.013}
\showDOI{\tempurl}


\bibitem[Saxena et~al\mbox{.}(2023)]%
        {saxena-etal-2023-vendorlink}
\bibfield{author}{\bibinfo{person}{Vageesh Saxena}, \bibinfo{person}{Nils Rethmeier}, \bibinfo{person}{Gijs van Dijck}, {and} \bibinfo{person}{Gerasimos Spanakis}.} \bibinfo{year}{2023}\natexlab{}.
\newblock \showarticletitle{{V}endor{L}ink: An {NLP} approach for Identifying {\&} Linking Vendor Migrants {\&} Potential Aliases on {D}arknet Markets}. In \bibinfo{booktitle}{\emph{Proceedings of the 61st Annual Meeting of the Association for Computational Linguistics (Volume 1: Long Papers)}}, \bibfield{editor}{\bibinfo{person}{Anna Rogers}, \bibinfo{person}{Jordan Boyd-Graber}, {and} \bibinfo{person}{Naoaki Okazaki}} (Eds.). \bibinfo{publisher}{Association for Computational Linguistics}, \bibinfo{address}{Toronto, Canada}, \bibinfo{pages}{8619--8639}.
\newblock
\urldef\tempurl%
\url{https://doi.org/10.18653/v1/2023.acl-long.481}
\showDOI{\tempurl}


\bibitem[Seyler et~al\mbox{.}(2021a)]%
        {darkjargon}
\bibfield{author}{\bibinfo{person}{Dominic Seyler}, \bibinfo{person}{Wei Liu}, \bibinfo{person}{XiaoFeng Wang}, {and} \bibinfo{person}{ChengXiang Zhai}.} \bibinfo{year}{2021}\natexlab{a}.
\newblock \showarticletitle{Towards Dark Jargon Interpretation in Underground Forums}. In \bibinfo{booktitle}{\emph{Advances in Information Retrieval: 43rd European Conference on IR Research, ECIR 2021, Virtual Event, March 28 – April 1, 2021, Proceedings, Part II}}. \bibinfo{publisher}{Springer-Verlag}, \bibinfo{address}{Berlin, Heidelberg}, \bibinfo{pages}{393–400}.
\newblock
\showISBNx{978-3-030-72239-5}
\urldef\tempurl%
\url{https://doi.org/10.1007/978-3-030-72240-1_40}
\showDOI{\tempurl}


\bibitem[Seyler et~al\mbox{.}(2021b)]%
        {darkjargon_net}
\bibfield{author}{\bibinfo{person}{Dominic Seyler}, \bibinfo{person}{Wei Liu}, \bibinfo{person}{Yunan Zhang}, \bibinfo{person}{XiaoFeng Wang}, {and} \bibinfo{person}{ChengXiang Zhai}.} \bibinfo{year}{2021}\natexlab{b}.
\newblock \showarticletitle{DarkJargon.Net: A Platform for Understanding Underground Conversation with Latent Meaning}. In \bibinfo{booktitle}{\emph{Proceedings of the 44th International ACM SIGIR Conference on Research and Development in Information Retrieval}} (Virtual Event, Canada) \emph{(\bibinfo{series}{SIGIR '21})}. \bibinfo{publisher}{Association for Computing Machinery}, \bibinfo{address}{New York, NY, USA}, \bibinfo{pages}{2526–2530}.
\newblock
\showISBNx{9781450380379}
\urldef\tempurl%
\url{https://doi.org/10.1145/3404835.3462801}
\showDOI{\tempurl}


\bibitem[Shan and Sankaranarayana(2020)]%
        {shan2020behavioral}
\bibfield{author}{\bibinfo{person}{Sylvester Shan} {and} \bibinfo{person}{Ramesh Sankaranarayana}.} \bibinfo{year}{2020}\natexlab{}.
\newblock \bibinfo{title}{Behavioral profiling of darknet marketplace vendors}.
\newblock
\newblock


\bibitem[Son et~al\mbox{.}(2023)]%
        {SoKModeration}
\bibfield{author}{\bibinfo{person}{Donghyun Son}, \bibinfo{person}{Byounggyu Lew}, \bibinfo{person}{Kwanghee Choi}, \bibinfo{person}{Yongsu Baek}, \bibinfo{person}{Seungwoo Choi}, \bibinfo{person}{Beomjun Shin}, \bibinfo{person}{Sungjoo Ha}, {and} \bibinfo{person}{Buru Chang}.} \bibinfo{year}{2023}\natexlab{}.
\newblock \showarticletitle{Reliable Decision from Multiple Subtasks through Threshold Optimization: Content Moderation in the Wild}. In \bibinfo{booktitle}{\emph{Proceedings of the Sixteenth ACM International Conference on Web Search and Data Mining}} (Singapore, Singapore) \emph{(\bibinfo{series}{WSDM '23})}. \bibinfo{publisher}{Association for Computing Machinery}, \bibinfo{address}{New York, NY, USA}, \bibinfo{pages}{285–293}.
\newblock
\showISBNx{9781450394079}
\urldef\tempurl%
\url{https://doi.org/10.1145/3539597.3570439}
\showDOI{\tempurl}


\bibitem[Soska and Christin(2015)]%
        {silkroad}
\bibfield{author}{\bibinfo{person}{Kyle Soska} {and} \bibinfo{person}{Nicolas Christin}.} \bibinfo{year}{2015}\natexlab{}.
\newblock \showarticletitle{Measuring the Longitudinal Evolution of the Online Anonymous Marketplace Ecosystem}. In \bibinfo{booktitle}{\emph{Proceedings of the 24th USENIX Conference on Security Symposium}} (Washington, D.C.) \emph{(\bibinfo{series}{SEC'15})}. \bibinfo{publisher}{USENIX Association}, \bibinfo{address}{USA}, \bibinfo{pages}{33–48}.
\newblock
\showISBNx{9781931971232}


\bibitem[Soussan and Kjellgren(2014)]%
        {NPS-forum}
\bibfield{author}{\bibinfo{person}{Christophe Soussan} {and} \bibinfo{person}{Anette Kjellgren}.} \bibinfo{year}{2014}\natexlab{}.
\newblock \showarticletitle{Harm reduction and knowledge exchange—a qualitative analysis of drug-related Internet discussion forums}.
\newblock \bibinfo{journal}{\emph{Harm reduction journal}} \bibinfo{volume}{11}, \bibinfo{number}{1} (\bibinfo{year}{2014}), \bibinfo{pages}{1--9}.
\newblock


\bibitem[Steiger et~al\mbox{.}(2021)]%
        {moderator_psych}
\bibfield{author}{\bibinfo{person}{Miriah Steiger}, \bibinfo{person}{Timir~J Bharucha}, \bibinfo{person}{Sukrit Venkatagiri}, \bibinfo{person}{Martin~J. Riedl}, {and} \bibinfo{person}{Matthew Lease}.} \bibinfo{year}{2021}\natexlab{}.
\newblock \showarticletitle{The Psychological Well-Being of Content Moderators: The Emotional Labor of Commercial Moderation and Avenues for Improving Support}. In \bibinfo{booktitle}{\emph{Proceedings of the 2021 CHI Conference on Human Factors in Computing Systems}} (Yokohama, Japan) \emph{(\bibinfo{series}{CHI '21})}. \bibinfo{publisher}{Association for Computing Machinery}, \bibinfo{address}{New York, NY, USA}, Article \bibinfo{articleno}{341}, \bibinfo{numpages}{14}~pages.
\newblock
\showISBNx{9781450380966}
\urldef\tempurl%
\url{https://doi.org/10.1145/3411764.3445092}
\showDOI{\tempurl}


\bibitem[Suntwal et~al\mbox{.}(2019)]%
        {fact_verif}
\bibfield{author}{\bibinfo{person}{Sandeep Suntwal}, \bibinfo{person}{Mithun Paul}, \bibinfo{person}{Rebecca Sharp}, {and} \bibinfo{person}{Mihai Surdeanu}.} \bibinfo{year}{2019}\natexlab{}.
\newblock \showarticletitle{On the Importance of Delexicalization for Fact Verification}. In \bibinfo{booktitle}{\emph{Proceedings of the 2019 Conference on Empirical Methods in Natural Language Processing and the 9th International Joint Conference on Natural Language Processing (EMNLP-IJCNLP)}}. \bibinfo{publisher}{Association for Computational Linguistics}, \bibinfo{address}{Hong Kong, China}, \bibinfo{pages}{3413--3418}.
\newblock
\urldef\tempurl%
\url{https://doi.org/10.18653/v1/D19-1340}
\showDOI{\tempurl}


\bibitem[Wiegand et~al\mbox{.}(2021)]%
        {abusive_implicit}
\bibfield{author}{\bibinfo{person}{Michael Wiegand}, \bibinfo{person}{Josef Ruppenhofer}, {and} \bibinfo{person}{Elisabeth Eder}.} \bibinfo{year}{2021}\natexlab{}.
\newblock \showarticletitle{Implicitly Abusive Language {--} What does it actually look like and why are we not getting there?}. In \bibinfo{booktitle}{\emph{Proceedings of the 2021 Conference of the North American Chapter of the Association for Computational Linguistics: Human Language Technologies}}. \bibinfo{publisher}{Association for Computational Linguistics}, \bibinfo{address}{Online}, \bibinfo{pages}{576--587}.
\newblock
\urldef\tempurl%
\url{https://doi.org/10.18653/v1/2021.naacl-main.48}
\showDOI{\tempurl}


\bibitem[Yang et~al\mbox{.}(2017)]%
        {klingon}
\bibfield{author}{\bibinfo{person}{Hao Yang}, \bibinfo{person}{Xiulin Ma}, \bibinfo{person}{Kun Du}, \bibinfo{person}{Zhou Li}, \bibinfo{person}{Haixin Duan}, \bibinfo{person}{Xiaodong Su}, \bibinfo{person}{Guang Liu}, \bibinfo{person}{Zhifeng Geng}, {and} \bibinfo{person}{Jianping Wu}.} \bibinfo{year}{2017}\natexlab{}.
\newblock \showarticletitle{How to Learn Klingon without a Dictionary: Detection and Measurement of Black Keywords Used by the Underground Economy}. In \bibinfo{booktitle}{\emph{2017 IEEE Symposium on Security and Privacy (SP)}}. \bibinfo{pages}{751--769}.
\newblock
\urldef\tempurl%
\url{https://doi.org/10.1109/SP.2017.11}
\showDOI{\tempurl}


\bibitem[Yang et~al\mbox{.}(2018)]%
        {distant_ner}
\bibfield{author}{\bibinfo{person}{Yaosheng Yang}, \bibinfo{person}{Wenliang Chen}, \bibinfo{person}{Zhenghua Li}, \bibinfo{person}{Zhengqiu He}, {and} \bibinfo{person}{Min Zhang}.} \bibinfo{year}{2018}\natexlab{}.
\newblock \showarticletitle{Distantly Supervised {NER} with Partial Annotation Learning and Reinforcement Learning}. In \bibinfo{booktitle}{\emph{Proceedings of the 27th International Conference on Computational Linguistics}}. \bibinfo{publisher}{Association for Computational Linguistics}, \bibinfo{address}{Santa Fe, New Mexico, USA}, \bibinfo{pages}{2159--2169}.
\newblock
\urldef\tempurl%
\url{https://aclanthology.org/C18-1183}
\showURL{%
\tempurl}


\bibitem[Yuan et~al\mbox{.}(2018)]%
        {cantreader}
\bibfield{author}{\bibinfo{person}{Kan Yuan}, \bibinfo{person}{Haoran Lu}, \bibinfo{person}{Xiaojing Liao}, {and} \bibinfo{person}{XiaoFeng Wang}.} \bibinfo{year}{2018}\natexlab{}.
\newblock \showarticletitle{Reading Thieves{\textquoteright} Cant: Automatically Identifying and Understanding Dark Jargons from Cybercrime Marketplaces}. In \bibinfo{booktitle}{\emph{27th USENIX Security Symposium (USENIX Security 18)}}. \bibinfo{publisher}{USENIX Association}, \bibinfo{address}{Baltimore, MD}, \bibinfo{pages}{1027--1041}.
\newblock
\showISBNx{978-1-939133-04-5}
\urldef\tempurl%
\url{https://www.usenix.org/conference/usenixsecurity18/presentation/yuan-kan}
\showURL{%
\tempurl}


\bibitem[Zannettou et~al\mbox{.}(2018)]%
        {zannettou2018gab}
\bibfield{author}{\bibinfo{person}{Savvas Zannettou}, \bibinfo{person}{Barry Bradlyn}, \bibinfo{person}{Emiliano De~Cristofaro}, \bibinfo{person}{Haewoon Kwak}, \bibinfo{person}{Michael Sirivianos}, \bibinfo{person}{Gianluca Stringini}, {and} \bibinfo{person}{Jeremy Blackburn}.} \bibinfo{year}{2018}\natexlab{}.
\newblock \showarticletitle{What is gab: A bastion of free speech or an alt-right echo chamber}. In \bibinfo{booktitle}{\emph{Companion Proceedings of the The Web Conference 2018}}. \bibinfo{pages}{1007--1014}.
\newblock


\bibitem[Zhang et~al\mbox{.}(2019)]%
        {zhang2019your}
\bibfield{author}{\bibinfo{person}{Yiming Zhang}, \bibinfo{person}{Yujie Fan}, \bibinfo{person}{Wei Song}, \bibinfo{person}{Shifu Hou}, \bibinfo{person}{Yanfang Ye}, \bibinfo{person}{Xin Li}, \bibinfo{person}{Liang Zhao}, \bibinfo{person}{Chuan Shi}, \bibinfo{person}{Jiabin Wang}, {and} \bibinfo{person}{Qi Xiong}.} \bibinfo{year}{2019}\natexlab{}.
\newblock \showarticletitle{Your style your identity: Leveraging writing and photography styles for drug trafficker identification in darknet markets over attributed heterogeneous information network}. In \bibinfo{booktitle}{\emph{The World Wide Web Conference}}. \bibinfo{pages}{3448--3454}.
\newblock


\bibitem[Zhang et~al\mbox{.}(2020)]%
        {zhang2020dstyle}
\bibfield{author}{\bibinfo{person}{Yiming Zhang}, \bibinfo{person}{Yiyue Qian}, \bibinfo{person}{Yujie Fan}, \bibinfo{person}{Yanfang Ye}, \bibinfo{person}{Xin Li}, \bibinfo{person}{Qi Xiong}, {and} \bibinfo{person}{Fudong Shao}.} \bibinfo{year}{2020}\natexlab{}.
\newblock \showarticletitle{dstyle-gan: Generative adversarial network based on writing and photography styles for drug identification in darknet markets}. In \bibinfo{booktitle}{\emph{Annual Computer Security Applications Conference}}. \bibinfo{pages}{669--680}.
\newblock


\bibitem[Zhao et~al\mbox{.}(2016)]%
        {china_ug}
\bibfield{author}{\bibinfo{person}{Kangzhi Zhao}, \bibinfo{person}{Yong Zhang}, \bibinfo{person}{Chunxiao Xing}, \bibinfo{person}{Weifeng Li}, {and} \bibinfo{person}{Hsinchun Chen}.} \bibinfo{year}{2016}\natexlab{}.
\newblock \showarticletitle{Chinese underground market jargon analysis based on unsupervised learning}. In \bibinfo{booktitle}{\emph{2016 IEEE Conference on Intelligence and Security Informatics (ISI)}}. \bibinfo{pages}{97--102}.
\newblock
\urldef\tempurl%
\url{https://doi.org/10.1109/ISI.2016.7745450}
\showDOI{\tempurl}


\bibitem[Zhu and Bhat(2021)]%
        {euph_mlm}
\bibfield{author}{\bibinfo{person}{Wanzheng Zhu} {and} \bibinfo{person}{Suma Bhat}.} \bibinfo{year}{2021}\natexlab{}.
\newblock \showarticletitle{Euphemistic Phrase Detection by Masked Language Model}. In \bibinfo{booktitle}{\emph{Findings of the Association for Computational Linguistics: EMNLP 2021}}. \bibinfo{publisher}{Association for Computational Linguistics}, \bibinfo{address}{Punta Cana, Dominican Republic}, \bibinfo{pages}{163--168}.
\newblock
\urldef\tempurl%
\url{https://doi.org/10.18653/v1/2021.findings-emnlp.16}
\showDOI{\tempurl}


\bibitem[Zhu et~al\mbox{.}(2021)]%
        {zhu2021self}
\bibfield{author}{\bibinfo{person}{Wanzheng Zhu}, \bibinfo{person}{Hongyu Gong}, \bibinfo{person}{Rohan Bansal}, \bibinfo{person}{Zachary Weinberg}, \bibinfo{person}{Nicolas Christin}, \bibinfo{person}{Giulia Fanti}, {and} \bibinfo{person}{Suma Bhat}.} \bibinfo{year}{2021}\natexlab{}.
\newblock \showarticletitle{{Self-supervised Euphemism Detection and Identification for Content Moderation}}. In \bibinfo{booktitle}{\emph{2021 IEEE Symposium on Security and Privacy (SP)}}. IEEE, \bibinfo{pages}{229--246}.
\newblock


\end{thebibliography}
\begin{appendices}
\section{Annotation guidelines.}
\label{appendix}

% We present some supplementary materials in this section. 
% % We provide an algorithm for masking and negative sampling process of training data preparation in algorithm~\ref{alg:data_preparation2}.

% \vspace{0.2cm}
% \noindent\textbf{Seed terms and extracted terms.}
% % \subsection{Word Lists}
% We provide a variety of word lists used for training \model{} and the results of jargon extraction for qualitative evaluation in Section~\ref{qaulitative_evaluation}.
% In Table~\ref{tab:drug_seed_terms}, we list the 46 drug seed terms used for applying our distantly supervised approach. These terms represent the known drug jargon from which we can train the model context of drug usage. 
% We also present the top 100 extracted drug jargon from each jargon extraction approach in Table~\ref{tab:top100_word_list}. 

% \vspace{0.2cm}
% \noindent\textbf{Annotation guidelines.}
% \subsection{Annotation Guidelines}
We provide the annotation guidelines developed after the pilot session. The detailed guidelines with examples are as follows.
\begin{itemize}
    \item Objective: Find words used as drug jargon in the given sentences.
    \vspace{0.1cm}
    \item When the word ``drug(s)'' is used to indicate an instance of the drug, annotate it as drug jargon. (e.g., I was high on \textbf{drugs} at the time.)
    \vspace{0.1cm}
    \item When the word ``drug(s)'' is used to describe just the concept of drugs, annotate it as NOT drug jargon. (e.g., I dismissed it as just a party \textbf{drugs}.)
    \vspace{0.1cm}
    \item When a word indicating the form of a drug is used as a substitute for drug words, annotate it as drug jargon. (e.g., The \textbf{bars} can get you pretty high.)
    \vspace{0.1cm}
    \item When a word indicating the form of a drug is used alongside drug words, annotate it as NOT drug jargon. (e.g., He had a bunch of Xanax \textbf{bars}.)
    \vspace{0.1cm}
    \item When a word refers to a legal drug, but the context is discussing it for abusive purposes, annotate it as drug jargon. (e.g., You can get a great stim high from \textbf{propylhexedrine} in otc inhalers.)
\end{itemize}

\begin{figure}[t]
    \centering
    \includegraphics[width=0.99\linewidth]{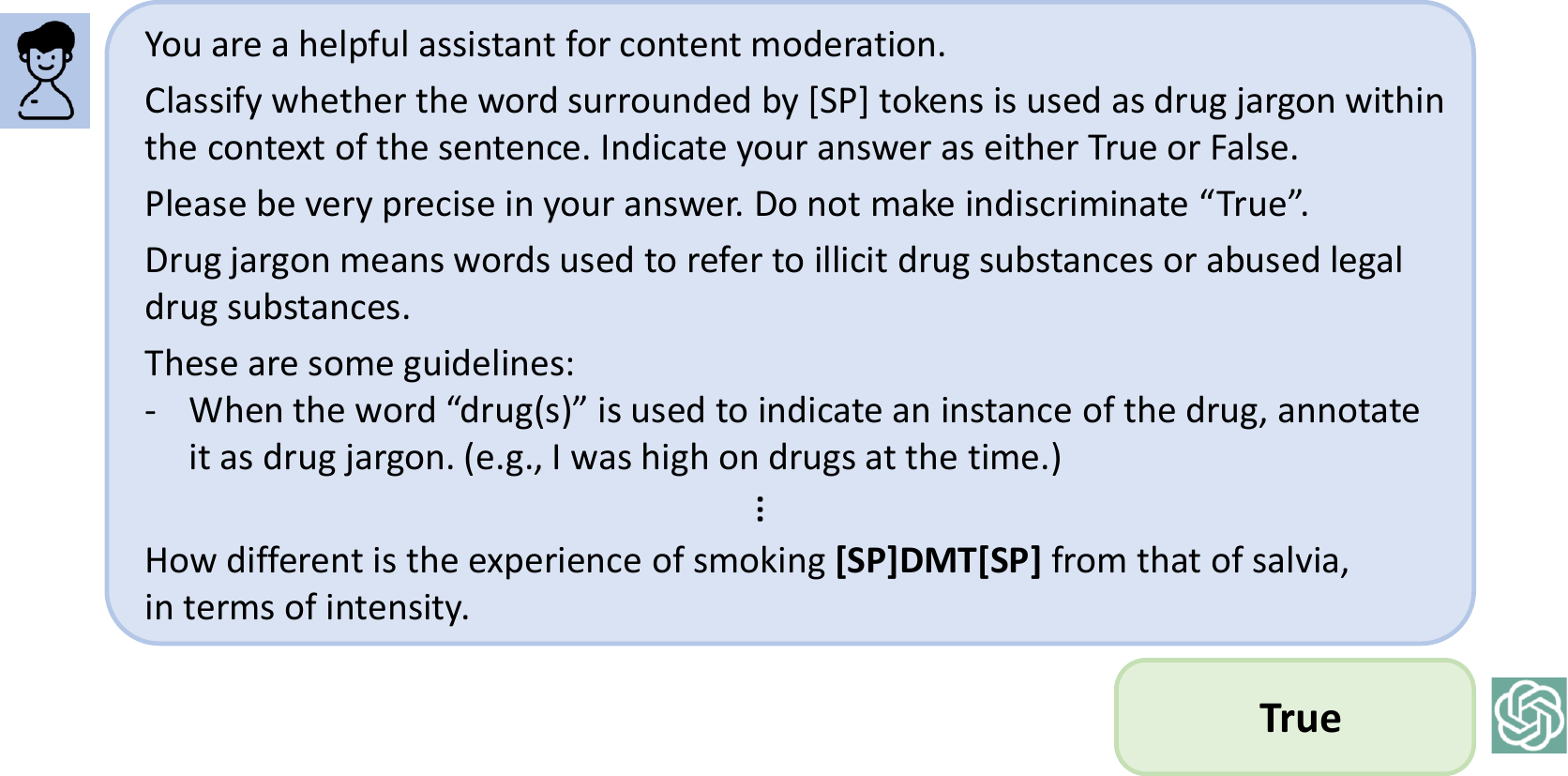}
    \caption{\diff{Prompts fed into GPT4o-mini. The definition of drug jargon in Section~\ref{preliminaries} and the annotation guidelines in Appendix~\ref{appendix} are used.}}
    \label{fig:prompts}
\end{figure}

\diff{
\section{Baselines}
\label{baselines}
For all relevant baseline models, hyperparameter $K$ was empirically tuned to achieve \textit{the best performance}.
\begin{itemize}[leftmargin=*]
    \item \textbf{Word2Vec}~\cite{mikolov2013efficient}
    Word2Vec learns vector representation of words. 
    We employ the CBOW mode, which trains the model by predicting the middle word based on its surrounding context.
    We use the most\_similar function of the Gensim library implementation of Word2Vec to obtain the most similar words to the 46 seed terms according to cosine similarity.
    This process generated jargon lists of 183 and 175 words from the Reddit Drug corpus and the Silk Road Forum corpus, respectively.
    \vspace{0.1cm}
    \item \textbf{CantReader}~\cite{cantreader}
    CantReader leverages the difference in word usage across different corpora (dark, legitimate, and reputable) to find euphemistic jargon.
    We use our forum corpora as the dark corpus to extract a drug jargon list. 
    The list is constructed by taking the top $K$ words produced by CantReader.
    We set $K$ as 120 for the Reddit Drug dataset and 220 for the Silk Road Forum dataset.
    \vspace{0.1cm}
    \item \textbf{Zhu et al.}~\cite{zhu2021self}
    A BERT-based MLM model is pretrained on the target corpus.
    Then, sentences containing drug seed terms are filtered to keep only sentences with informative drug contexts.
    From the informative sentences, the \textit{drug seed terms} are masked.
    The MLM model predicts words that can replace the mask tokens.
    The system produces a jargon list by sorting accumulated MLM probabilities.
    It is capable of finding both euphemistic and non-euphemistic drug jargon.
    We take top $K$ words to create the banlist.
    We set $K$ as 40 and 50 for each dataset, respectively.
    \vspace{0.1cm}
    \item \textbf{PETD}~\cite{lee-etal-2022-searching}
    PETDetection finds potentially euphemistic terms using a combination of distributional similarities and sentiment-based metrics. 
    It involves extracting phrase candidates from the sentence, filtering them based on their relevance to the topic (i.e., \textit{drug}), paraphrasing to identify sentiment shifts, and ranking the phrases. 
    For each sentence, we mark a word as the drug jargon when it passes all the stages (extraction, filtering, paraphrasing, and ranking).
    \vspace{0.1cm}
    \item \textbf{MLM \textit{(w/o pretrain})} 
    We newly present another baseline that uses MLM probability in the different manner with Zhu et al.
    Instead of masking known drug words, we mask the \textit{target word} to be inspected.
    If a drug seed term appears in the top $K$ words predicted by the MLM, the masked word is predicted as a drug jargon in the sentence.
    We use the BERT-base-uncased as the MLM model, which has been pretrained on generic web texts.
    We set $K$ as 100 for both datasets.
    \vspace{0.1cm}
    \item \textbf{MLM}
    The same approach as above, but we further pretrain the MLM model on the target corpus.
    % Since BERT-base was trained on general text corpora, it may not be suitable for understanding drug contexts.
    % Therefore, we further pretrain a BERT-based MLM model to perform the MLM task on the target corpus.
    % Pretraining is done with the MLM task on the target corpus.
    We set $K$ as 50 for the Reddit Drug dataset and 30 for the Silk Road Forum dataset.
    \vspace{0.1cm}
    \item \textbf{DarkBERT}~\cite{jin2023darkbert}
    DarkBERT is a RoBERTa-based language model pretrained on Dark Web data. Leveraging the domain-specific knowledge acquired from Dark Web data, it demonstrates superior performance in detecting threat keywords, including illicit drug terms. Following the methodology outlined in the original paper~\cite{jin2023darkbert}, if a drug seed term appears in the top $K$ related words predicted by the MLM, the masked word is identified as drug jargon in the sentence. 
    We set $K$ to 40 for both datasets.
    \vspace{0.1cm}
    \item \textbf{GPT4o-mini}~\cite{gpt4omini}
    GPT4o-mini can be utilized for content moderation through prompt learning. As an assistant to content moderators, we use specifically designed prompts (as shown in Figure~\ref{fig:prompts}) and feed them with sentences where the inspected words are highlighted with [SP] tokens. 
    This baseline demonstrates the utility of recent large language models and prompt learning in the context of drug content moderation.

    \vspace{0.1cm}
    \item \textbf{\model{} ablation variations}  Settings of \model{} without using negative data from STMS (\textit{w/o} $Neg_{STMS}$), pretraining BERT on the target domain (\textit{w/o pretrain}), word attribute module (\textit{w/o word}), or both pretraining and word attribute module (\textit{w/o pretrain\&word}).
    These baselines suggest the importance of information obtained from each component for drug jargon detection.
\end{itemize}
}

\section{Experimental Settings}
\label{exp_settings}
Evaluations were done on a machine with two Intel(R) Xeon(R) Silver 4214R CPU @ 2.40GHz and two NVIDIA GeForce RTX 4090s.
For the BERT models used in \textit{Classification} modules, we used the bert-base-uncased model with the embedding dimension of 768, max length of 128, batch size of 32, learning rate of 1e-5, and warmup ratio of 0.1.
For further pretraining, we ran run\_mlm.py on HuggingFace with max sequence length of 512, validation split percentage of 10, batch size of 32 train epochs of 3.
Then, we trained the classification models with the AdamW optimizer and cross entropy loss for a maximum of 10 epochs with early stopping enabled. We used Python version 3.10 for all implementations.

We also train the baseline models as follows:
For Word2Vec baselines, min count of 10, vector size of 100, window size of 10, epochs of 10 are used as parameters.
For Zhu et al.'s approach, we follow the parameters described in the original paper.
For the MLM baselines, we ran run\_mlm.py on HuggingFace using bert-base-uncased with max sequence length of 512, validation split percentage of 10 batch size of 32 train epochs of 3.
\diff{For PETD, we use the topic word as \textit{drug}~\footnote{\diff{We also experimented with using our 46 drug seed terms as topic words, but this resulted in poorer performance.}} and follow the other parameters described in the original paper.
}

\section{Ethical Considerations}
\label{ethical}
Because our work is closely related to user-generated content, we consider the ethical concerns during our study. First, we use only publicly released data accessible to anyone, as mentioned in Section~\ref{dataset}. Other existing works use the same data for clustering drug vendors in darknet markets~\cite{manolache2022veridark, saxena-etal-2023-vendorlink, shan2020behavioral,zhang2020dstyle, zhang2019your} or finding drug jargon in social media~\cite{euph_mlm, zhu2021self}. Second, we anonymized or removed the information that could possibly lead to any harm, such as usernames, email addresses, and PGP keys.
Finally, we will open our annotated datasets to only researchers who fill out a request form with clear research purposes and verifiable identities.
We believe that the contributions of our work on automatic content moderation and the application of NLP techniques could outweigh the potential risks derived from our study.

\section{How to address potential subjectivity and bias in annotation}
\label{potential_sub_bias}
We constructed evaluation datasets through manual annotation. 
However, this process may introduce potential subjectivity or bias from the annotators, which could affect the quality of the constructed dataset. 
To mitigate these effects, we implemented the following process:
First, we recruited two expert annotators from a cybersecurity company specializing in monitoring malicious online activities. They possess substantial knowledge of drug jargon and high expertise in related tasks.
Second, as described in Section~\ref{annotation}, we conducted a pilot session using the first 100 sentences of each dataset. Subsequently, we developed detailed annotation guidelines with examples, as outlined in Appendix~\ref{appendix}. 
As shown in Table~\ref{tab:annotation_statistics}, this resulted in a very high inter-annotator agreement. 
Finally, in cases of disagreement between annotators, thorough discussions were held to reach a consensus.
Through these steps, we made careful efforts to minimize potential bias and subjectivity.

\section{Cross Domain Experiments}
\label{cross_domain}
To investigate whether \model{} can perform well in unseen domains, we conduct jargon detection using the model trained on the other dataset. The experimental results are shown in Table~\ref{tab:cross_domain}. In both datasets, \model{} trained on the target dataset produced higher performance. However, the cross-domain models also produced reasonably high performance comparable to the pretrained MLM approach. This suggests that the languages of the two datasets do differ, but knowledge learned from one domain could be transferred to the target domain. The results suggest that while training the classification model on the target domain remains ideal, our approach has the ability to generalize to unseen domains that do not have sufficient training data (such as emerging social media sites).
\begin{table}[t]
    \centering
    \ra{1.3}
    \caption{\diff{Evaluation results on cross domain performance. $\normalsize{\text{T}}_{\text{A} \rightarrow \text{B}}$ denotes performance of \model{} trained on corpus A and evaluated on corpus B.}}
    % \\ Reddit* and SR1* on the left most column refer to the model trained on Reddit and SR1 corpus, respectively.
    \footnotesize
    % \resizebox{0.99\linewidth}{!}{%
        \begin{tabular} {l c c c}
            \toprule
            \textbf{Domain Transfer} & \textbf{Precision}  & \textbf{Recall}   & \textbf{F1-score} \\
            \midrule
            % \vspace{0.05cm}
            $\normalsize{\text{T}}_{\text{\scriptsize{Reddit}} \rightarrow \text{\scriptsize{Reddit}}}$ &  \diff{0.6659} &  \diff{0.6197} &  \diff{0.6419} \\
            $\normalsize{\text{T}}_{\text{\scriptsize{SR}} \rightarrow \text{\scriptsize{Reddit}}}$ & \diff{0.5868} & \diff{0.5973} & \diff{0.5920} \\
            % \vspace{0.05cm}
            $\normalsize{\text{T}}_{\text{\scriptsize{SR}} \rightarrow \text{\scriptsize{SR}}}$ & \diff{0.5805} & \diff{0.6934} & \diff{0.6320} \\
            % \vspace{0.05cm}
            $\normalsize{\text{T}}_{\text{\scriptsize{Reddit}} \rightarrow \text{\scriptsize{SR}}}$ & \diff{0.5616} & \diff{0.5497} &  \diff{0.5556} \\

            \bottomrule
        \end{tabular}%
    % }
    \label{tab:cross_domain}
\end{table}

\section{Application to Other Fields}
\label{other_fields}
Content moderation can encompass other types of forbidden content, such as discussion on weapons, sexually explicit, or hate speech. While \model{} was designed for drug jargon detection, its foundational principles permit broader applications. 
However, the success of \model{} rests on the distinctiveness of jargon contexts from non-jargon. Illicit drug terms have specific linguistic features and contextual cues that might not be as evident in other domains. For \model{} to be effectively repurposed for other areas, the target jargon should have contexts as distinguishable as drug terms. 
Thus, while the potential exists for \model{}'s adaptation to domains with distinct linguistic characteristics, its efficacy in such extensions warrants rigorous future exploration.

\vspace{0.1cm}
\noindent\textbf{Pilot experiment on sexually explicit domain.}
To evaluate the potential for adapting \model{} to other domains, we conducted a pilot study focusing on the moderation of sexually explicit discussions. 
The training corpus comprised 1,000,000 sentences extracted from user discussions on the GAB platform~\cite{zannettou2018gab}, with 43 sexual seed words, which were manually identified, serving as seed terms. 
Given the absence of the annotated dataset for evaluation, an additional set of 10 sexual words was used as test terms to assess \model{}'s performance in detecting these terms within the test corpus.
The outcomes demonstrated a precision of \textit{0.6585}, a recall of \textit{0.5562}, and an F1-score of \textit{0.6030}~\footnote{The experimental results do not reflect precise accuracy due to the test set's construction, which relied solely on the presence of test terms without annotation.}. 
These results indicate that \model{} holds promise for application in domains beyond its original design, even when not optimized through hyperparameter adjustments.

\end{appendices}

\end{document}